\pdfoutput=1

\documentclass[11pt]{article}

\usepackage{EMNLP2022}
\usepackage{EMNLP2022}

\usepackage{times}
\usepackage{latexsym}

\usepackage{tablefootnote}

\usepackage[T1]{fontenc}

\usepackage[utf8]{inputenc}

\usepackage{microtype}

\usepackage{inconsolata}

\usepackage{times}
\usepackage{latexsym}
\usepackage[T1]{fontenc}
\usepackage[utf8]{inputenc}
\usepackage{microtype}
\usepackage{amsmath,amsthm,amssymb}
\usepackage{enumitem}
\usepackage{algorithmic}
\usepackage{mathtools}
\usepackage{multirow}
\usepackage{booktabs}
\usepackage{graphicx}
\usepackage{caption}
\usepackage{subcaption}
\usepackage{float}
\usepackage{dsfont}
\usepackage{array}
\usepackage{color}
\usepackage[ruled, lined, linesnumbered, commentsnumbered, longend]{algorithm2e}

\usepackage{pifont}

\mathchardef\mhyphen="2D 

\newcommand{\dataset}{{\fontfamily{bch}\selectfont{\textsc{RobustLR}}}\xspace}
\newcommand{\equivalence}{Logical Equivalence}
\newcommand{\contrast}{Logical Contrast}

\newcommand{\conjcs}{\textsc{C-CS}}
\newcommand{\disjcs}{\textsc{D-CS}}
\newcommand{\negcs}{\textsc{N-CS}}
\newcommand{\contraes}{\textsc{C-ES}}
\newcommand{\distaes}{\textsc{D1-ES}}
\newcommand{\distbes}{\textsc{D2-ES}}

\newcommand{\roberta}{RoBERTa-Large}

\DeclareMathAlphabet{\mathcal}{OMS}{cmsy}{m}{n}
\SetMathAlphabet{\mathcal}{bold}{OMS}{cmsy}{b}{n}

\newcolumntype{P}[1]{>{\centering\arraybackslash}p{#1}}

\title{\dataset{}: A Diagnostic Benchmark for Evaluating Logical Robustness of Deductive Reasoners}

\author{ 
Soumya Sanyal \quad
Zeyi Liao \quad 
Xiang Ren \\
{University of Southern California} \\
{\texttt{\{soumyasa, zeyiliao, xiangren\}@usc.edu}}
}

\begin{document}
\maketitle
\begin{abstract}
	Transformers have been shown to be able to perform deductive reasoning on inputs containing rules and statements written in English natural language. However, it is unclear if these models indeed follow rigorous logical reasoning to arrive at the  prediction, or rely on spurious correlation patterns in making decision. A strong deductive reasoning model should consistently understand the semantics of different logical operators. To this end, we present \dataset{}, a deductive reasoning-based diagnostic benchmark that evaluates the robustness of language models to minimal logical edits in the inputs and different logical equivalence conditions. In our experiments with RoBERTa, T5, and GPT3, we show that the models trained on deductive reasoning datasets with various logical operations do not perform consistently on the \dataset{} test set, thus showing that the models are not robust to our proposed logical perturbations. Further, we observe that the models find it especially hard to learn logical negation operator. Our results demonstrate the shortcomings of current language models in logical reasoning, and call for the development of better inductive biases to teach the logical semantics to language models. All the datasets and code base have been made publicly available. \footnote{\url{https://github.com/INK-USC/RobustLR}}
\end{abstract}

\section{Introduction}
Building systems that can automatically reason over a given context to generate valid logical inferences is a long pursued goal within the field of AI \citep{Mccarthy1959ProgramsWC,Rocktschel2017EndtoendDP,Manhaeve2019DeepProbLogNP}. Recently, \citet{ruletaker} have shown that language models \cite{liu2019roberta,raffel2019exploring} are able to emulate deductive reasoning on a logical rulebase (\textit{theory}) containing rules and declarative statements written in natural language. 
While this is impressive, it is unclear if these models are able to perform logical reasoning robustly by understanding the semantics of the logical operators and the different logical conditions involving such operators.

\begin{figure}[t]
\vspace{-0.2cm}
	\centering
	\includegraphics[width=\columnwidth]{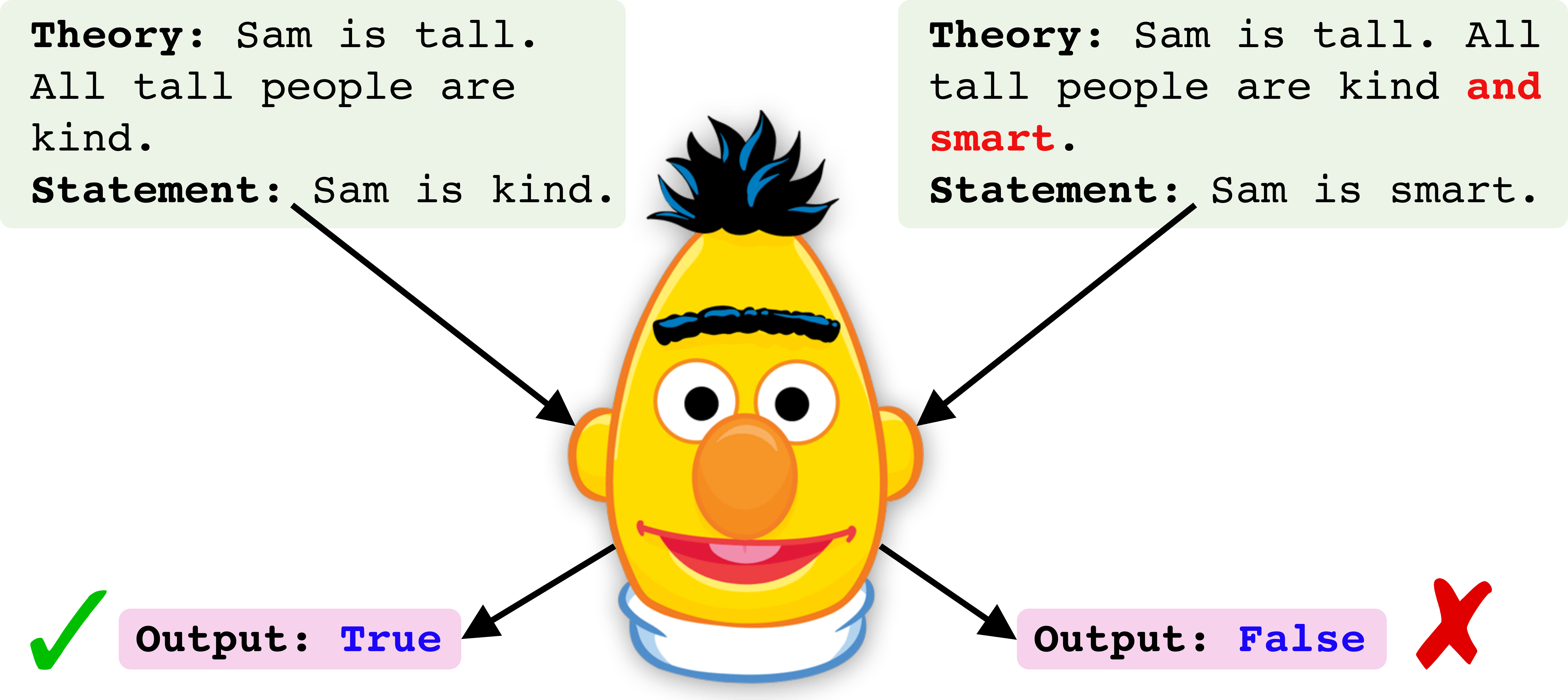}
	\caption{\small \textbf{Overview of \dataset{}.} We expect a strong deductive reasoning model should be robust to logical variations in the input. Here, the model fails to understand the logical conjunction in second example and predicts the wrong entailment of the statement.}
	\label{fig:motivation}
	\vspace{-0.3cm}
\end{figure}

Logical reasoning is an important skill required in various NLP tasks such as NLI \citep{nli_dataset}, Question Answering \citep{yang-etal-2018-hotpotqa}, Multi-turn Dialogue Reasoning \cite{cui-etal-2020-mutual}, etc. Models used to solve such tasks may use spurious patterns to reach to the predictions, rather than following the intended logical reasoning process. Additionally, these models might only understand certain ways of expressing the inference knowledge (e.g., rules) and not possess systematic generalization \cite{Gontier2020MeasuringSG}. Hence, it is important to ensure that language models can consistently use the logical operators when described in natural language. Prior works \cite{gururangan-etal-2018-annotation,chen-durrett-2019-understanding,mccoy-etal-2019-right} have found that models solving different reasoning tasks tend to exploit spurious correlations between the context/question and the label. But logical reasoning needs special considerations as there are very well-defined relationships on how different logical operators modify any given context. Hence, it is important to understand if models use these logical relationships consistently to solve a task. To the best of our knowledge, a study evaluating a language model's logical consistency on different logical operations is currently missing.

A key desirable property of a strong deductive reasoning model is \textit{logical robustness}. This is the ability to make consistent predictions on inputs that have some logical modifications. In Figure \ref{fig:motivation}, we show how the lack of logical robustness can lead to wrong inferences in a model. Thus, to test this, we develop \dataset{}, a diagnostic benchmark for evaluating logical robustness across two main aspects. First, we aim to evaluate how robust these models are when tested on the three logical operators: conjunction ($\land$), disjunction ($\lor$), and negation ($\neg$). Inspired by the idea of contrast sets \cite{gardner-etal-2020-contrast-set}, we design the \contrast{} set, where theories are minimally modified so that we can test the model's robustness across logical operators. Examples of this are shown in Figure \ref{fig:example}(b) and \ref{fig:example}(c). Next, we study the model's ability of reasoning consistently across \textit{logical paraphrases}. A logical paraphrase uses equivalence conditions in logic to replace a rule with another equivalent form, essentially rewriting the surface form of the rule. This poses a different challenge than the more common language paraphrases such as synonym changes, voice modifications, style changes, etc., because the model needs to understand that the underlying logical structure of the two paraphrased sentences mean the same thing.
An example of the equivalence perturbation is shown in Figure \ref{fig:example}(d). Based on this, we design 
the \equivalence{} set containing  three logical equivalences.

In this work, we study three language models: RoBERTa \cite{liu2019roberta}, T5 \cite{raffel2019exploring}, and GPT-3 \cite{gpt3}. To test the model performance on \dataset{}, we first fine-tune them on deductive reasoning training datasets containing the logical operators and then evaluate on the \dataset{} test sets. Overall, we find that language models (LMs) fine-tuned on different deductive reasoning datasets are not sufficiently robust to the \contrast{} and \equivalence{} sets. Specifically, we find that models are more inconsistent with logical negations in sentences. We also find that using larger models such as T5-11B improves the performance to an extent, but they still perform worse compared to human performance on \dataset{}. We show that it is partly due to spurious correlations in the data and the inherent difficulty of the task. Thus, we use \dataset{} to demonstrate some key limitations of the language models trained for deductive reasoning. We hope that this research will encourage as a test bed to evaluate robustness of deductive reasoning models.

\vspace{-0.1cm}
\section{Deductive Reasoning}
\vspace{-0.1cm}

In deductive reasoning, we predict whether a given theory $T$ supports a statement $s$ or not. We define a theory $T$ as a set of facts \begin{small}$F=\{f_1, f_2, \dots, f_n \}$\end{small} and rules \begin{small}$R=\{r_1, r_2, \dots, r_m\}$\end{small} expressed in natural language (See Figure \ref{fig:example} for an example). For a given theory, a statement can be either provably supported, provably unsupported (i.e., the negation of the statement is provable), or not provable at all. This is a 3-class classification problem, with the labels \textit{True}, \textit{False}, and \textit{Unknown}, respectively. In this work, we focus on this task, where we expect the model to correctly predict the entailment of a statement for a given theory. In Figure \ref{fig:example}(a), \ref{fig:example}(c), and \ref{fig:example}(d), the statement is entailed by the theory, leading to the label True while in Figure \ref{fig:example}(b), the statement is not provable given the facts and rules. It can be proved by simply using fact $f_1$ and rule $r_1$ to derive the statement. Formally, we define the proof set of a statement $s$, denoted by $G(T, s)$, as the set of rules and facts that are required to obtain the statement $s$ from the theory.

\section{Evaluating LMs for Logical Robustness}

\vspace{-0.1cm}
\subsection{Logical Robustness}
\label{sec:overview}
\vspace{-0.1cm}
We consider a deductive reasoner (language model) to be \textit{logically robust} if the model behavior is consistent across various logical perturbations, as illustrated in Figure \ref{fig:motivation}. Specifically, we evaluate logical robustness on two types of perturbations.

\paragraph{Logical contrastive edits} Here, we test the model's ability to correctly capture the semantics of different logical operators, when presented in minimally edited contrast inputs. A contrast set \cite{gardner-etal-2020-contrast-set} is one where the input is changed minimally, but meaningfully, such that there is (typically) some change in label. These probes test the LM's robustness to conjunction ($\land$), disjunction ($\lor$), and negation ($\neg$).

\vspace{-0.1cm}
\paragraph{Logical paraphrases} Here, we evaluate whether the model performs consistently when shown the same input with different \textit{logical paraphrases}. A theory can be logically paraphrased by modifying the rules using standard logical equivalence conditions. \footnote{\url{https://en.wikipedia.org/wiki/Logical_equivalence}} These probes evaluate the model's consistency in solving logically equivalent theories, when logical conditions are used to rewrite the input.

A strong deductive reasoning model should be robust to both the minimally edited contrast inputs and logical paraphrases. Overall, these evaluation sets probe a deductive reasoning model to check whether it indeed learns the semantics of the logical operators and their underlying working principles.

\vspace{-0.1cm}
\subsection{Notations}
\label{sec:notations}
\vspace{-0.1cm}

In this work, we consider two predicate forms: unary and binary. A unary predicate contains one argument and is denoted by $X(a)$. Similarly, a binary predicate is represented as  $X(a, b)$. Here, $X$ is the predicate relation and $a, b$ are the variables. An \textit{atomic} predicate is defined as either a predicate or the negation of the predicate (denoted as $\neg X(a)$). A \textit{complex} predicate can contain multiple predicates (or their negated forms) combined using logical operators conjunction and disjunction.

\begin{figure}[t]
\vspace{-0.5cm}
	\centering
	\includegraphics[width=0.7\columnwidth]{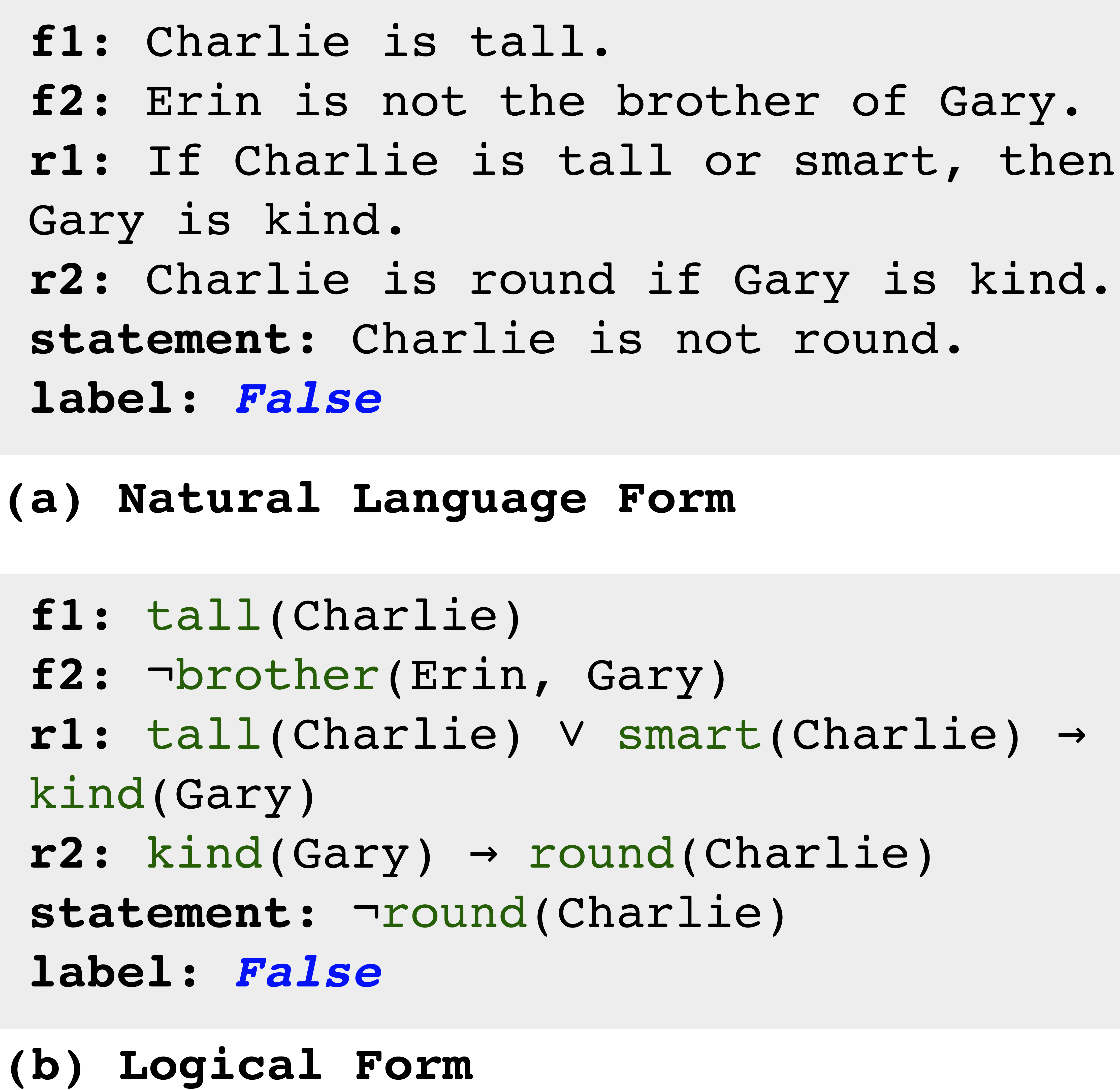}
	\caption{\small \textbf{Logical Form of a Theory.} (a) A theory in natural language. (b). The corresponding logical form of the theory. Refer to Section \ref{sec:notations} for more details.}
	\label{fig:logical_form}
\vspace{-0.3cm}
\end{figure}

Internally, we maintain a symbolic representation of these facts and rules, enabling us to later create the different evaluation sets of \dataset{}. A fact is symbolically represented by a predicate. In this work, we consider all facts as atomic predicates. A rule is symbolically represented by a logical connection between predicates, separated by the ``\textit{implies that}'' logical symbol ($\implies$). Thus, a rule can be defined as $p \implies q$, where the LHS $p$ and RHS $q$ are atomic or complex predicates. If both $p$ and $q$ consist of atomic predicates, then the rule is called a \textit{simple} rule. A \textit{compound} rule is one where $p$ and/or $q$ contain some complex predicates connected by the conjunction or disjunction operator. An example of a natural language theory and its corresponding logical form is shown in Figure \ref{fig:logical_form}. Here, facts $f_1$ and $f_2$ are unary atomic and binary predicates, respectively. Rule $r_1$ is a compound rule, with the LHS $p$ of the rule being a complex predicate. Rule $r_2$ is a simple rule.

\begin{table}[t]
\vspace{-0.5cm}
	\centering
	\resizebox{0.85\columnwidth}{!}{%
		\begin{tabular}{lcccc}
			\toprule
			\textbf{Modified Rule} & \textbf{Facts}	& \textbf{Statement}	& \textbf{Label}    & \textbf{Group}	\\
			\midrule
			$p \implies q$ & $\{p\}$ & $q$  & \textit{True} & \textsc{base} \\
			\midrule
			$p \land t \implies q$ & $\{p\}$ & $q$  & \textit{Unknown}  & \textsc{conj} \\
			$p \land t \implies q$ & $\{p, t\}$ & $q$  & \textit{True}  & \textsc{conj} \\
			\midrule
			$p \land t \implies q$ & $\{p, \neg t\}$ & $q$  & \textit{Unknown}  & \textsc{conj+neg} \\
			$p \land t \implies \neg q$ & $\{p\}$ & $q$  & \textit{Unknown} & \textsc{conj+neg} \\
			$p \land t \implies \neg q$ & $\{p, t\}$ & $q$  & \textit{False}    & \textsc{conj+neg} \\
			$p \land t \implies \neg q$ & $\{p, \neg t\}$ & $q$  & \textit{Unknown} & \textsc{conj+neg} \\
			\bottomrule
		\end{tabular}%
	}
	\caption{\label{tab:conj_contrast_list} \small \textbf{Conjunction Contrast Perturbations}. The minimal edits done to a base theory (first row) for testing the conjunction and negation reasoning abilities. The group reflects the overall change in theory w.r.t. the base theory.}
\vspace{-0.2cm}
\end{table}

\vspace{-0.1cm}
\subsection{\contrast{} Sets}
\label{sec:contrast_sets}
In this evaluation set, we probe the ability of the model to robustly understand the three different logical operators ($\land, \lor, \neg$). For this, we develop different contrast sets \citep{gardner-etal-2020-contrast-set} with minimal editing of the theory, probing specific reasoning abilities of different operators. The key intuition is to evaluate if the model is able to understand the minor changes in the theory brought by the addition of logical operators, and predict the change in label accordingly.

\begin{figure}[t]
	\centering
	\includegraphics[width=\columnwidth]{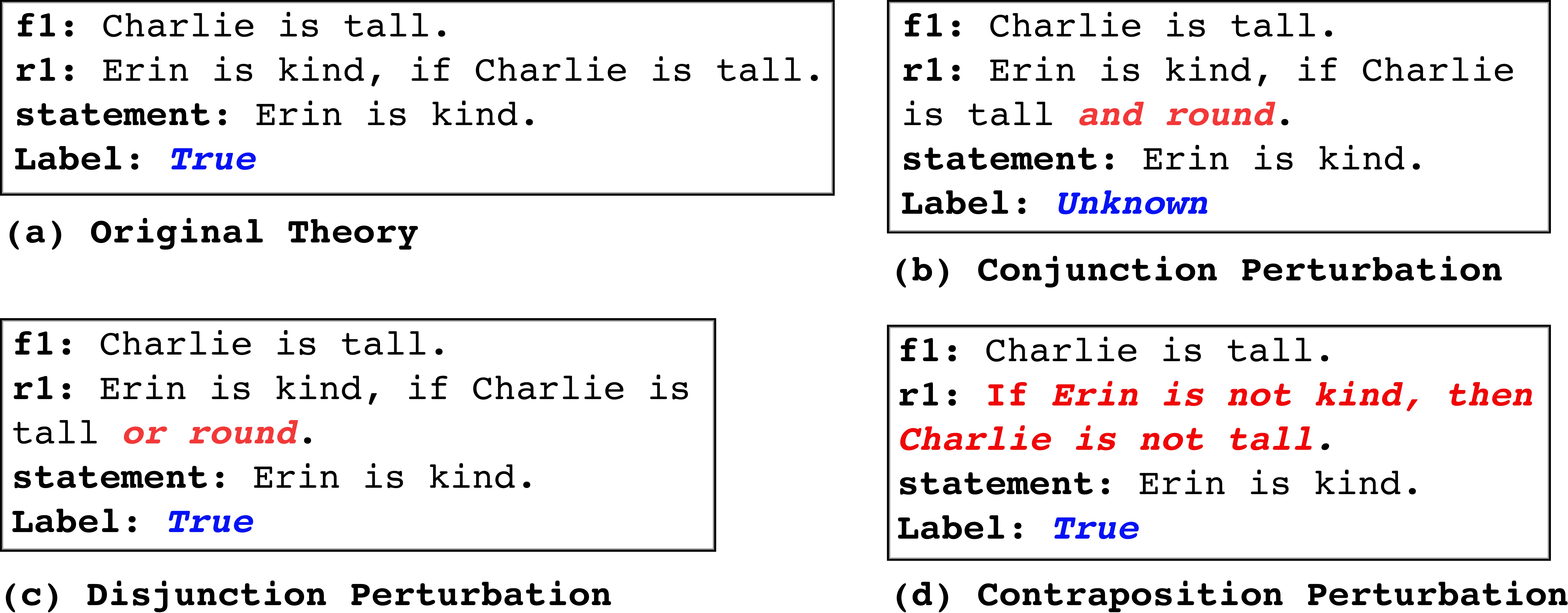}
	\caption{\textbf{Examples of perturbations in \dataset{}.} (a) An original theory contains facts, rules, a statement, and the entailment label. The \contrast{} set perturbations using conjunction and disjunction are shown in bold in (b) and (c), respectively. In (d), we show one of the \equivalence{} perturbations where the rule is paraphrased using logical contraposition. Please refer to Sections \ref{sec:contrast_sets} and \ref{sec:equivalence_sets} for more details.}
	\label{fig:example}
\end{figure}

For a given theory $T$ and statement $s$, we first select a rule to be modified such that it is part of the proof set $G(T, s)$. This ensures that our perturbation would influence the model's reasoning process while predicting entailment of the statement $s$. Next, we add an unseen predicate $t$ to the rule LHS $p$ of one of the rules using conjunction ($\land$) or disjunction ($\lor$). In some further variants of perturbations, we include the predicate $t$ (or the negated $\neg t$) as a fact in the theory, leading to different labels. Lastly, we also negate the rule RHS $q$ to introduce the logical negation ($\neg$) perturbations. Based on the logical operator in the perturbation, we broadly divide the \contrast{} set into three types: Conjunction Contrast Set (\conjcs{}), Disjunction Contrast Set (\disjcs{}), and Negation Contrast Set (\negcs{}). Examples of these perturbations are shown in Figure \ref{fig:example}. The perturbations for \conjcs{} are listed in Table \ref{tab:conj_contrast_list}. Please refer to Appendix \ref{app:contrast_perturbations} for the other perturbations. We categorize these perturbations into groups based on the logical operators involved in the perturbation with respect to the base theory. E.g., the three groups for \conjcs{} are \textsc{base}, \textsc{conj}, \textsc{conj+neg}, as shown in Table \ref{tab:conj_contrast_list}. If a model performs accurately on the \contrast{} set, we expect that the model understands the semantics of the logical operators robustly.

\paragraph{Evaluation Protocol}
To evaluate the logical robustness to contrast perturbations, we first finetune the language model on a deductive reasoning dataset containing different combination of logical operators. Then, we report the model performance on these evaluation datasets as the weighted-F1 score from the Scikit-learn \cite{scikit-learn}. The weighted-F1 score modifies the macro-F1 to take any label imbalance into account. For instance, we have label imbalance by design of the perturbations in the \contrast{} sets shown in Tables \ref{tab:conj_contrast_list}, \ref{tab:disj_contrast_list}, and \ref{tab:neg_contrast_list}. We use the model's prediction for the base theory and all its perturbations to compute the F1-score at a theory level, and then average this score across all theories in the evaluation set.

\subsection{\equivalence{} Sets}
\label{sec:equivalence_sets}
The \equivalence{} set contain theories where the underlying symbolic representation of a rule is replaced by another representation that is logically equivalent. The logical equivalent form of a rule can be derived from standard logical equivalence conditions, as defined below:
\begin{itemize}[leftmargin=*]
	\small
	\item \textbf{Contrapositive}: $p \implies q \equiv \neg q \implies \neg p$
	\item \textbf{Distributive 1}: $(p \implies q) \land (p \implies r) \equiv p \implies (q \land r)$
	\item \textbf{Distributive 2}: $(p \implies q) \land (r \implies q) \equiv (p \lor r) \implies q$
\end{itemize}
Here $p, q, r$ can be both atomic predicates or complex predicates. Based on the above conditions, the \equivalence{} set is divided into three types: Contrapositive Equivalence Set (\contraes{}), Distributive 1 Equivalence Set (\distaes{}), and Distributive 2 Equivalence Set (\distbes{}). For the (\contraes{}) set, every rule $r_i$ in the theory $T$ is replaced by the logically equivalent form to create a new logically equivalent theory $T'$. An example of this perturbation is shown in Figure \ref{fig:example} (d). Similarly, for the \distaes{} and \distbes{} sets, a pair of rules in $T$ are merged according to the equivalence to create a new theory $T'$.

In both instances, the theory $T'$ still has the same label for a given statement, as the logical steps required to solve the task remains the same. These modifications are more challenging than traditional surface-level paraphrases of the natural language text, as it requires the model to understand the equivalence of different symbolic representations.

\paragraph{Evaluation Protocol}
Similar to the \contrast{} set evaluation, we finetune a language model on a deductive reasoning dataset and report the weighted-F1 score for the base theory and the corresponding logical paraphrase, averaged across all theories in the evaluation set.

\section{The \dataset{} Dataset}
In this section, we describe details about the \dataset{} dataset domains, sampling, and filtering procedure.

\subsection{Dataset Domain}
\label{sec:dataset_domain}

\paragraph{Facts} The domains of the predicate relation $X$ and variable $a$ in the unary predicate $X(a)$ are the simple English adjectives and the proper names, respectively. Examples of this predicate form are ``green(Alex)'', ``kind(John)'', etc. Each predicate is associated to the English template sentence form ``\{$a$\} is \{$X$\}.''. For the binary predicate $X(a, b)$, we consider family relationships and proper names as the domain of $X$ and $a$ respectively. Some examples of this predicate form are ``daughter(Mary, Gary)'',``father(Bob, John)'', etc. Each predicate is associated with template sentences such as ``\{$a$\} is the \{$X$\} of \{$b$\}.'', ``The \{$X$\} of \{$b$\} is \{$a$\}.'', etc. Note that, currently, we do not enforce any gender constraints on the names, thus allowing predicates such as ``daughter(Bob, Gary)'', which might be unlikely based on the genders associated statistically to names in English.

\paragraph{Rules} For the rules, we follow the same domain as mentioned above for facts. We allow rules containing unary predicates, binary predicates, or a combination of both. Examples of some simple rules consisting of atomic predicates are ``green(Alex) $\implies$ daughter(Bob, Gary)'', ``$\neg $ father(Bob, John) $\implies$ kind(John)'', etc. Similarly, examples of some compound rules containing complex predicates are ``green(Alex) $\lor$ smart(Bob) $\implies$ daughter(Bob, Gary) $\land$ $\neg$ kind(John)'', etc. We note that, for the sake of keeping the theories deterministic, we do not allow the disjunction operator in the RHS of a compound rule. A rule of the form $p \implies q$ is associated with templates such as ``If \{$p$\} then \{$q$\}.'', ``\{$q$\} if \{$p$\}.'', etc., where the $p$'s and $q$'s can be recursively resolved to their own templates as defined in the predicates.

\subsection{Dataset Sampling}
For sampling the theories in \dataset{}, we use a modified version of the Label-Priority sampling \cite{sampling_algorithm}. The detailed algorithm is described in Algorithm \ref{algo:sampling} in Appendix. At a high-level, we sample different predicates from the set of templates and assign the value $0$ or $1$ to them. After that, we divide the predicate set into multiple levels. This helps us in sampling theories with multi-hop reasoning depths. After that, rules are derived by connecting predicates with the same label between two different levels. Finally, the predicates with value $1$ at the $0^{th}$ level form the facts in the theory, the connections denote the rules, and the predicates in the last level denote some candidate statements.

\subsection{Filtering Statistical Features}
\label{sec:bias_filter}
In a contemporary work, \citet{sampling_algorithm} find that LMs are specifically prone to pick up any existing \textit{statistical features} that can be present in the training datasets. These are described as certain statistic of an instance that has a strong correlation with the label. Examples of statistical features are \#facts, \#rules, \#negation op, \#facts with negation, etc.

We introduce some changes in our sampling algorithm to minimize the influence of such statistical features. We define a check to ensure that the number of statements with the different labels are similar for any given theory in the training dataset. Since we have negations in the dataset, it allows us to exactly control the \textit{True} and \textit{False} label distribution per theory instance systematically. We control the \textit{Unknown} label by oversampling theories and discarding ones with skewed label distribution. Please refer to Appendix \ref{app:statistical_features} for more details.

%

\begin{table*}[t]
	\centering
	\resizebox{0.8\textwidth}{!}{%
		\begin{tabular}{lcccccccccccc}
			\toprule
			\multirow{2}{*}{\textbf{Data}}		& \multicolumn{4}{c}{\textbf{\roberta{}}}	& \multicolumn{4}{c}{\textbf{T5-Large}} & \multicolumn{4}{c}{\textbf{T5-3B}}	\\
			\cmidrule(r){2-5} \cmidrule(r){6-9} \cmidrule(r){10-13}
			& \textbf{Avg} & \conjcs{} & \disjcs{} & \negcs{} & \textbf{Avg} & \conjcs{} & \disjcs{} & \negcs{} & \textbf{Avg} & \conjcs{} & \disjcs{} & \negcs{} \\
			\midrule
			\texttt{NOT}        & 0.39 & 0.39 & 0.45 & 0.34 & 0.44 & 0.36 & 0.55 & 0.41 & 0.59 & 0.58 & 0.60 & 0.60 \\
			\texttt{AND+NOT}    & 0.52 & 0.56 & 0.53 & 0.48 & 0.47 & 0.43 & 0.55 & 0.42 & 0.57 & 0.58 & 0.57 & 0.55 \\
			\texttt{OR+NOT}     & 0.47 & 0.39 & 0.61 & 0.42 & 0.46 & 0.36 & 0.60 & 0.43 & 0.56 & 0.44 & 0.67 & 0.57 \\
			\texttt{All}        & 0.47 & 0.44 & 0.61 & 0.37 & 0.46 & 0.37 & 0.61 & 0.40 & 0.58 & 0.54 & 0.65 & 0.54 \\
			\bottomrule
		\end{tabular}%
	}
	\caption{\label{tab:main_results_contrast} Performance of \roberta{}, T5-Large, and T5-3B on \contrast{} sets. We report the weighted-F1 score for each subset, and average that for the \textbf{Avg} column. Please refer to Section \ref{sec:contrast_results} for more details.}
\end{table*}

\begin{table}[t]
	\centering
	\resizebox{0.88\columnwidth}{!}{%
		\begin{tabular}{lcccc}
			\toprule
			\textbf{Training Dataset} & \textbf{\roberta{}} & \textbf{T5-Large} & \textbf{T5-3B} & \textbf{T5-11B}\\
			\midrule
			\texttt{NOT}&1.00 &	1.00 &	0.98 & -  \\ 
			\texttt{AND+NOT}&1.00 &	0.99 &	0.97 & - \\
			\texttt{OR+NOT}&1.00 &	0.99 &	0.97 & - \\
			\texttt{All}&1.00 &	0.99 &	0.92 & 0.94 \\
			\bottomrule
		\end{tabular}%
	}
	\caption{\label{tab:results_iid} Performance of \roberta{}, T5-Large, T5-3B, and T5-11B on in-domain held-out set. Please refer to Section \ref{sec:results_in_domain} for more details.}
\end{table}

\section{Experimental Setup}
\label{sec:setup}

\paragraph{Training Data Details}
We use four different training datasets to fine-tune baselines, described as follows:
\textbf{\texttt{NOT}}: In this dataset, we allow negations in both facts and rules, but restrict to only using simple rules. Note that it is not possible to create a dataset without any operators (i.e., no negation, conjunction, and disjunction) as it would not be possible to have the \textit{False} label in that dataset.
\textbf{\texttt{AND+NOT}}: Here, we restrict the connector for compound rules to \texttt{AND} ($\land$). As before, we allow negations in facts and rules.
\textbf{\texttt{OR+NOT}}: Similar to \texttt{AND+NOT}, we restrict the connector of the compound rules to \texttt{OR} ($\lor$). We allow negations in facts and rules as before.
\textbf{\texttt{All}}: This dataset has all the three logical operators (\texttt{AND}, \texttt{OR}, \texttt{NOT}) present.

The dataset \texttt{All} contains all the logical operators we consider in the \contrast{} set. Thus, instances from the \texttt{All} dataset cover all forms of rules seen in these test sets. We aim to understand the effect of these training datasets on the evaluation sets by fine-tuning the model on each dataset separately. Please refer to Appendix \ref{app:dataset_statistics} for more details on the training and evaluation data statistics.

\paragraph{Models and Experiment Details}
Following prior works \cite{ruletaker,proofwriter,sanyal2022fairr}, we evaluate the performance of three language models: RoBERTa \cite{liu2019roberta} finetuned on the RACE dataset \cite{race}, T5 \cite{raffel2019exploring}, and GPT-3 \cite{gpt3}. Specifically, we evaluate the model checkpoints \roberta{}, T5-Large, T5-3B, T5-11B, and GPT-3. To evaluate a model, we first fine-tune it on one of the training dataset mentioned above, and then evaluate on the \contrast{} (Section \ref{sec:contrast_results}) and \equivalence{} (Section \ref{sec:equivalence_results}) evaluation sets. For T5-11B, we only finetune it on the \texttt{All} dataset due to compute constraints. For GPT-3, we evaluate its performance on a subset of the test sets using demonstrations. Please refer to Appendix \ref{app:implementation_details} for details on the input formats for each model and Appendix \ref{app:hyperparameters} for the hyperparameter settings and other implementation details.
\section{Results}


\subsection{In-domain Performance}
\label{sec:results_in_domain}
The performance of the LMs on the in-distribution held-out data are shown in Table \ref{tab:results_iid}. We note that the models are able to solve the in-distribution test dataset almost perfectly in all cases. This either means the model understands the logical reasoning task perfectly (which is unlikely) or it learns some spurious features to solve the task using shortcuts. Now, we evaluate these models on our test sets to check the logical robustness.

\subsection{Performance on \contrast{} set}
\label{sec:contrast_results}
\paragraph{Overall Result} We finetune \roberta{}, T5-Large, and T5-3B models on different training datasets and evaluate them on the three types of \contrast{} set. The results are shown in Table  \ref{tab:main_results_contrast}. Based on the \textbf{Avg} performance, we find that the models perform significantly worse on the \contrast{} set, compared to the almost perfect performance on the in-distribution test sets in Table \ref{tab:results_iid}. This shows that even after finetuning on the logical deductive reasoning datasets, these models do not learn the semantics of the logical operators in a robust manner, but likely use spurious correlations.

Additionally, we find that on average, model performance is similar for different training datasets, except for \texttt{NOT} dataset. While this result seems a bit surprising at first because we expect different training datasets to have varying effect on the performance, but computing the performance breakdown by perturbation type across different datasets reveals an interesting trend. We observe that training on related operators as the perturbation in the evaluation subset usually leads to better performance. For instance, we find that models trained on the \texttt{OR+NOT} training data perform better on the \disjcs{} compared to when trained on \texttt{NOT} or \texttt{AND+NOT} training data. Taking \roberta{} as an example, we observe that its performance on  \disjcs{} is 0.61 when finetuned on \texttt{OR+NOT} and is apparently greater than the performance on other contrast sets. A similar trend is observed for \conjcs{}. This is intuitive as training on the related operators helps the model to understand the semantics of the logic relatively better than just relying on having seen these at pre-training time. This shows that the model indeed requires some training data that is strongly aligned with the operators being evaluated the test set.
\begin{figure}[t]
\vspace{-0.3cm}
	\centering
	\includegraphics[width=0.85\columnwidth]{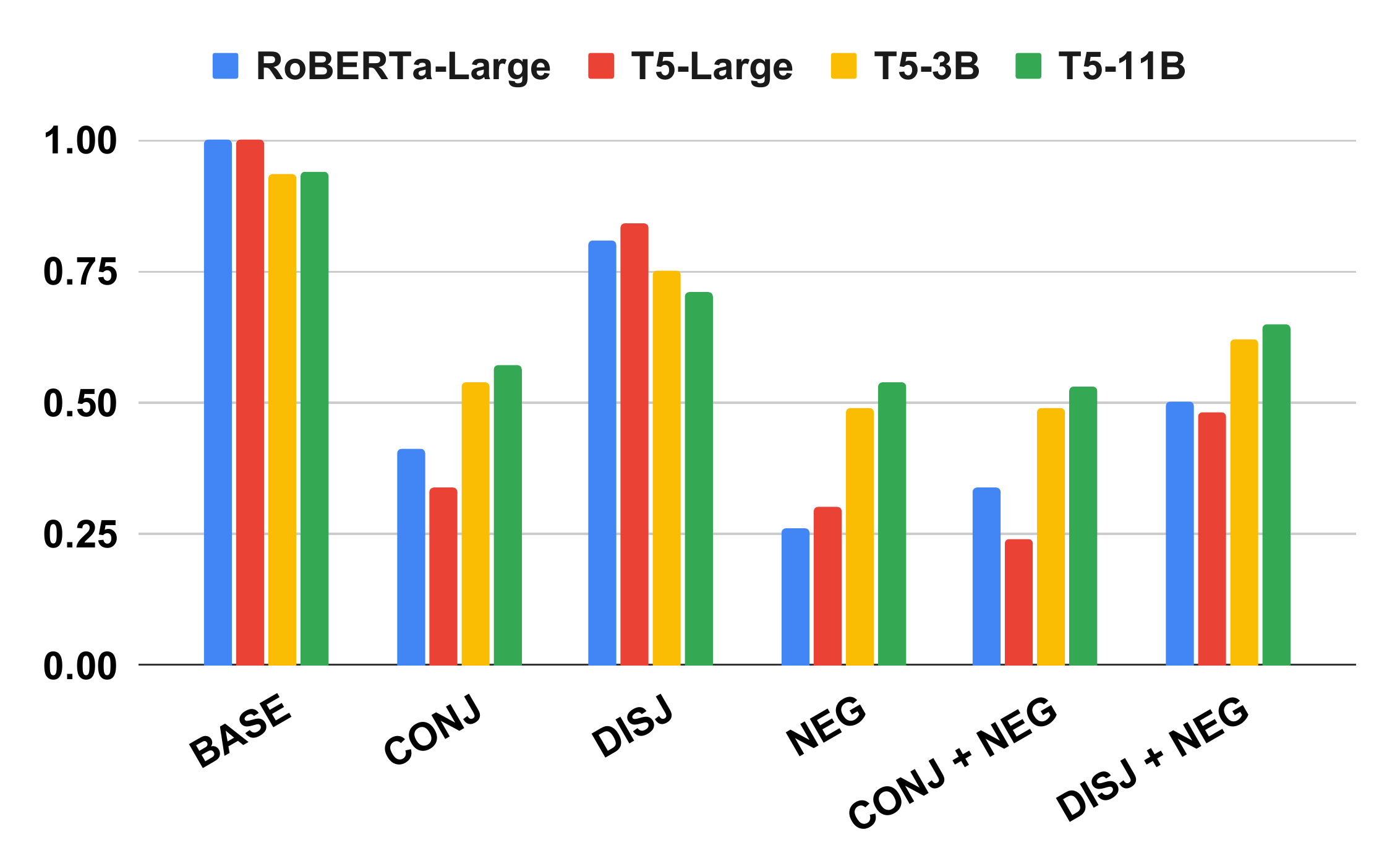}
	\caption{Performance comparison of \roberta{}, T5-Large, T5-3B, and T5-11B across different groups of contrast perturbations. Negations are hardest to learn across all settings. Refer to Section \ref{sec:contrast_results} for more details.}
	\label{fig:breakdown1}
	\vspace{-0.3cm}
\end{figure}

\begin{table*}[t]
	\centering
	\resizebox{0.88\textwidth}{!}{%
		\begin{tabular}{lcccccccccccc}
			\toprule
			\multirow{2}{*}{\textbf{Training Dataset}}		& \multicolumn{4}{c}{\textbf{\roberta{}}}	& \multicolumn{4}{c}{\textbf{T5-Large}} & \multicolumn{4}{c}{\textbf{T5-3B}}	\\
			\cmidrule(r){2-5} \cmidrule(r){6-9} \cmidrule{10-13}
			& \textbf{Avg} & \contraes{} & \distaes{} & \distbes{} & \textbf{Avg} & \contraes{} & \distaes{} & \distbes{} & \textbf{Avg} & \contraes{} & \distaes{} & \distbes{}   \\
			\midrule
			\texttt{NOT}        & 0.81 & 0.79 & 0.91 & 0.74 & 0.88 & 0.76 & 0.90 & 0.97 & 0.70 & 0.78 & 0.79 & 0.54 \\
			\texttt{AND+NOT}    & 0.87 & 0.80 & 0.88 & 0.94 & 0.88 & 0.77 & 0.90 & 0.96 & 0.83 & 0.76 & 0.81 & 0.93 \\
			\texttt{OR+NOT}     & 0.87 & 0.78 & 0.87 & 0.95 & 0.86 & 0.76 & 0.86 & 0.97 & 0.86 & 0.77 & 0.82 & 0.98 \\
			\texttt{All}        & 0.89 & 0.79 & 0.93 & 0.94 & 0.89 & 0.77 & 0.93 & 0.98 & 0.84 & 0.72 & 0.83 & 0.98 \\
			\bottomrule
		\end{tabular}%
	}
	\caption{\label{tab:main_results_equivalence} Performance of \roberta{}, T5-Large, and T5-3B on \equivalence{} sets. We report the weighted-F1 score, and average that for the \textbf{Avg} column. Please refer to Section \ref{sec:equivalence_results} for more details.}
\end{table*}

\paragraph{Variation with logical operators} Next, we want to understand which among the three operators are more challenging for the models to learn. To better understand this, we evaluate the models after finetuning on the \texttt{All} dataset and plot the model performance for different perturbation groups in Figure \ref{fig:breakdown1}. These groups (defined in Section \ref{sec:contrast_sets}) contain perturbations of a specific operator, as suggested by their names. We find that the most challenging operator is negation. This is evident from the lowest scores on the \textsc{neg} perturbation group among \textsc{conj}, \textsc{disj}, and \textsc{neg}. Further, this is also observed from the fact that performance generally drops when negation perturbations are introduced along with any other perturbations. For instance, we see an average drop of around 25\% between \textsc{disj} and \textsc{disj+neg}. This demonstrates that the model not able to learn the negation semantics very well. Lastly, we find that models find conjunction relatively harder than disjunction. Please refer to Appendix \ref{app:contrast_set_breakdown} for more details.

\subsection{Performance on \equivalence{} set}
\label{sec:equivalence_results}
\paragraph{Results on Contrapositive Equivalence} Next, we evaluate the fine-tuned LMs on the \equivalence{} sets. In Table \ref{tab:main_results_equivalence}, we observe that the model performance degrades by approximately 20\% for the \contraes{}, compared to the in-distribution performance in Table \ref{tab:results_iid}. Contraposition involves changing the rule into a format that has two negations, thus testing the limits of the model on understanding negations. From the experiments on \contrast{} sets, we know that negations are not well understood by the model. Thus, these results reinforce our previous findings. We do not find any significant changes in performance when trained on different operators. Thus, we conclude that, for this test set, it is not sufficient to just understand the semantics of the logical operators, but it rather requires a higher order understanding about the interactions between the logical operators and implications. Including such knowledge in deductive reasoning models is an interesting direction for future works.

\begin{table}[t]
	\centering
	\resizebox{0.95\columnwidth}{!}{%
		\begin{tabular}{lcccccccc}
			\toprule
			\multirow{2}{*}{\textbf{Models}}		& \multicolumn{4}{c}{\textbf{\contrast{} Set}}	& \multicolumn{4}{c}{\textbf{\equivalence{} Set}} \\
			\cmidrule(r){2-5} \cmidrule(r){6-9}
			& \textbf{Avg} & \conjcs{} & \disjcs{} & \negcs{} & \textbf{Avg} & \contraes{} & \distaes{} & \distbes{} \\
			\midrule
			From scratch    & 0.14 & 0.10 & 0.21 & 0.10 & 0.45 & 0.33 & 0.50 & 0.51 \\
			RoBERTa         & 0.47 & 0.44 & 0.61 & 0.37 & 0.76 & 0.40 & 0.93 & 0.94 \\
			T5-Large        & 0.46 & 0.37 & 0.61 & 0.40 & 0.89 & 0.77 & 0.93 & 0.98 \\
			T5-3B           & 0.58 & 0.54 & 0.65 & 0.54 & 0.84 & 0.72 & 0.83 & 0.98 \\
			T5-11B \tablefootnote{Fintuned for two epochs on the \texttt{All} training set.}    & 0.61 & 0.57 & 0.67 & 0.58 & 0.83 & 0.76 & 0.76 & 0.97 \\
			GPT-3 \tablefootnote{Evaluated on 500 samples for each test subset.}    & 0.36 & 0.34 & 0.50 & 0.25 & 0.67 & 0.36 & 0.78 & 0.87 \\
			\midrule
			human           & 0.88 & 0.87 & 0.94 & 0.84 & 0.91 & 0.81 & 0.97 & 0.96 \\

			\bottomrule
		\end{tabular}%
	}
	\caption{\label{tab:results_human_eval} Comparisons between  training a model from scratch, finetuning a pre-trained checkpoint at scale (RoBERTa, T5-Large, T5-3B, and T5-11B), using in-context learning (GPT-3), and human performance, on the \contrast{} and \equivalence{} sets. Please refer to Sections \ref{sec:human_evaluation} and \ref{sec:analysis} for more details.}
	\vspace{-0.3cm}
\end{table}

\paragraph{Results on Distributive Equivalence} For \distaes{} and \distbes{} test sets, we see a higher performance compared to \contraes{} as these equivalence conditions are relatively easier than the contraposition rule. This is because the distributive rules are similar to having compound rules in the dataset (that is already present in \texttt{AND+NOT}, \texttt{OR+NOT}, and \texttt{All} datasets). Between the two sets, we observe that \distaes{} is more challenging for the model. This indicates that models find conjunction operator harder than the disjunction, similar to our observations from Figure \ref{fig:breakdown1}. One reason for this can be the strictness involved in the conjunction operation. A rule with conjunction is true only if the individual parts are independently true.

\subsection{Human Evaluation}
\label{sec:human_evaluation}
To better understand the upper limit of \dataset{} evaluation sets, we ask 3 Computer Science graduate students to annotate 30 randomly sampled theories from each subset of \contrast{} and \equivalence{} sets. The results are shown in the last row of Table \ref{tab:results_human_eval}. We find that humans are significantly better than other baselines, performing around $27$\% higher on the \contrast{} set compared to T5-11B on average. Additionally, we see similar trends that humans find negation based perturbations (\negcs{}) hardest, followed by conjunction (\conjcs{}), and disjunction (\disjcs{}). Please refer to Appendix \ref{app:iaa_result} for further details.

\subsection{Analysis}
\label{sec:analysis}
\vspace{-0.0cm}

\paragraph{Performance of Larger LMs}
Here we evaluate the performance of some larger models on \dataset{}. The goal is to estimate the possible gains with scaling to large LMs. For this, we finetune a T5-11B using the \texttt{All} training dataset for two epochs. Additionally, we evaluate GPT3 \cite{gpt3} on a subset of 500 samples per test set, using demonstration-based in-context learning. The results are shown rows 5-6 in Table \ref{tab:results_human_eval}. We observe that increasing the model size from T5-Large to T5-11B indeed leads to a significant performance gain on most datasets. But it is still quite far compared to human performance. This suggests that scaling can potentially help with learning robust logical operations to an extent. Additionally, we find that GPT-3 is not able to perform well on these test sets. We hypothesize this is likely due to the mismatch of the training distribution of the GPT-3 model, versus our synthetic datasets, as we do not finetune the GPT-3 model on our training sets.

\paragraph{Effect of LM Pre-training}
In this part, we evaluate the usefulness of using a pre-trained checkpoint in our experiments, in comparison with training a \roberta{} architecture from scratch. We train both \roberta{} pre-trained checkpoint and a similar model from scratch using the \texttt{All} dataset, and evaluate on the \dataset{} sets. The results are shown in rows 1-2 in Table \ref{tab:results_human_eval}. We observe a significant drop in performance, demonstrating that knowledge learned during pre-training is crucial for this task.

\begin{figure}[t]
\vspace{-0.3cm}
	\centering
	\includegraphics[width=0.7\columnwidth]{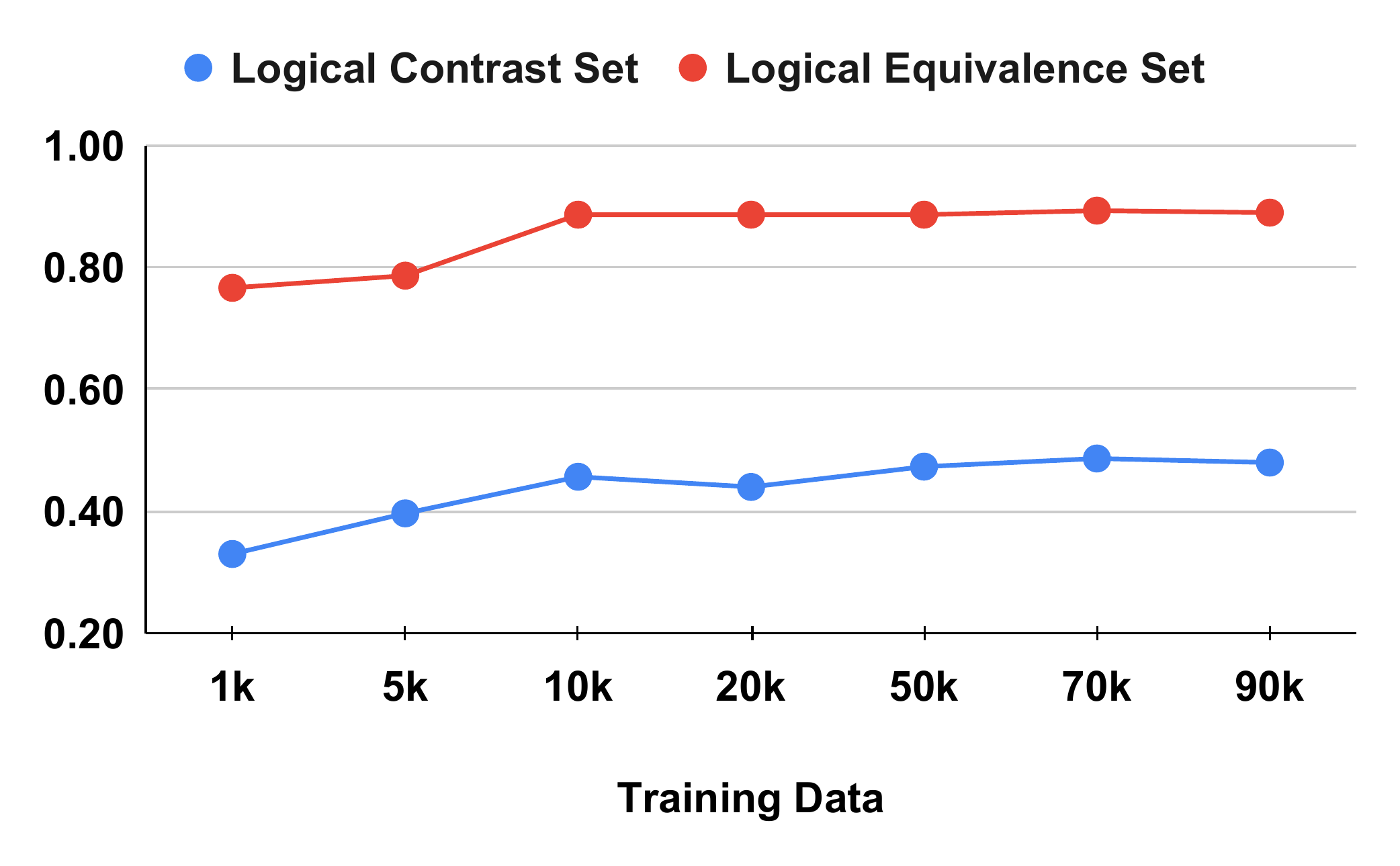}
	\caption{Average performance of \roberta{} on \contrast{} and \equivalence{} sets when trained on varying amount of \texttt{All} dataset.}
	\label{fig:train_data_variation}
	\vspace{-0.3cm}
\end{figure}

\paragraph{Effect of size of training data}
Next, in Figure \ref{fig:train_data_variation}, we plot the overall performance of \roberta{} model finetuned on a varying amount of \texttt{All} training data. We observe that the model performance increases with increasing amount of training data, as expected, and then saturates at a fixed level. This shows that there is no significant effect of using larger training datasets on model performance. We use 50k training samples in all datasets.

\begin{figure}[t]
	\centering
	\includegraphics[width=0.8\columnwidth]{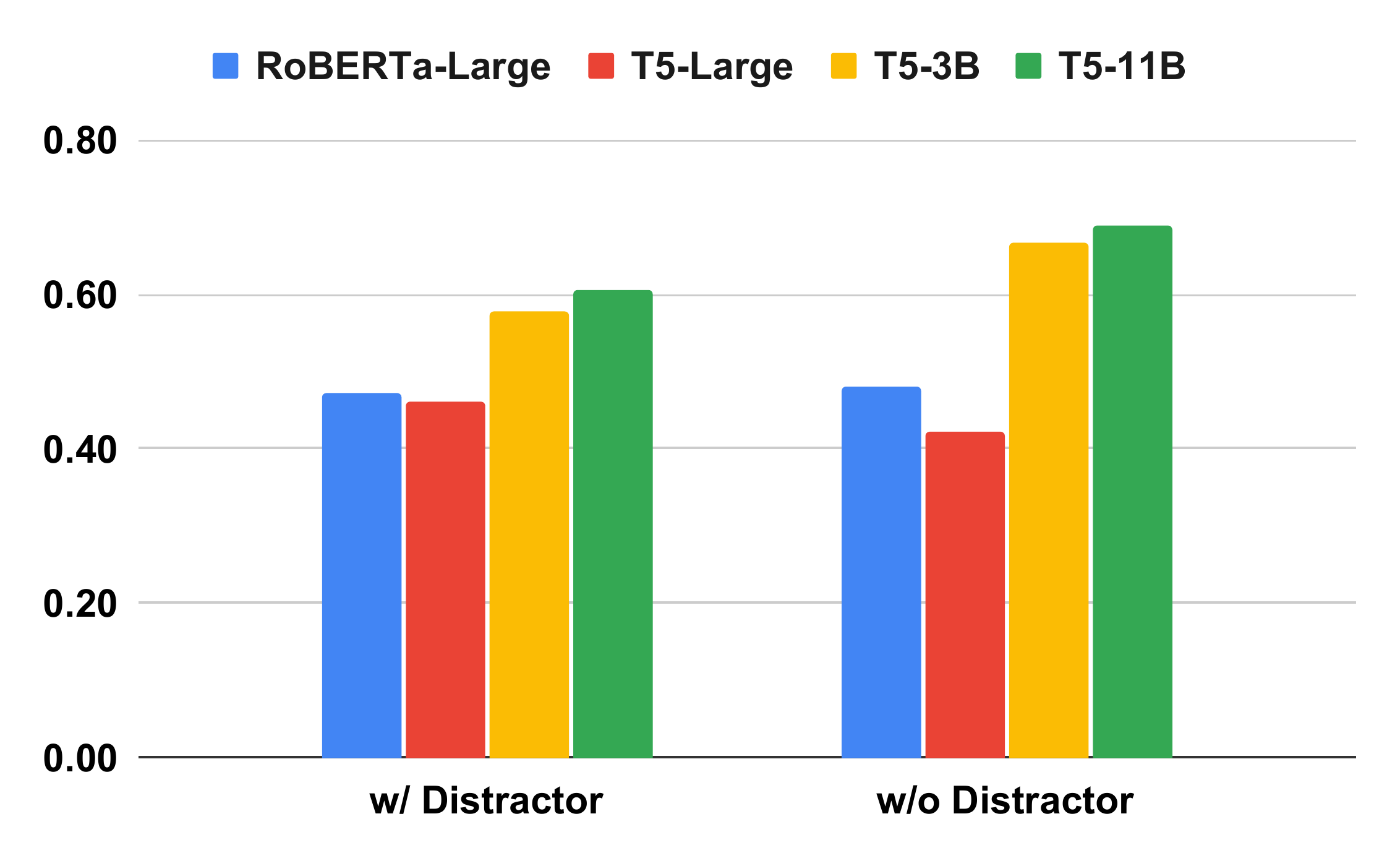}
	\caption{Performance comparison of \roberta{}, T5-Large, T5-3B, and T5-11B for two variants of \contrast{} sets: with and without distractors. Refer to Section \ref{sec:analysis} for more details.}
	\label{fig:distractor}
	\vspace{-0.3cm}
\end{figure}

\paragraph{Effect of distractors}
We define distractors as the facts and rules that are not part of the proof set for a given theory and statement.
Figure \ref{fig:distractor} depicts the effect of distractors on the model performance, when finetuned on the \texttt{All} dataset and evaluated on the \contrast{} set. We observe that the performance generally improves (or stays similar) on the variant without distractors. This shows that retrieving the relevant facts and rules in the theory from a given set of sentences is a non-trivial challenge. Thus, performing both retrieval and entailment prediction in a single model can lead to some performance degradation.

\begin{table}[t]
	\centering
	\resizebox{0.48\textwidth}{!}{%
		\begin{tabular}{lcccccccc}
			\toprule
			\multirow{2}{*}{\texttt{all} Dataset}		& \multicolumn{4}{c}{\textbf{\contrast{} Set}}	& \multicolumn{4}{c}{\textbf{\equivalence{} Set}} \\
			\cmidrule(r){2-5} \cmidrule(r){6-9}
			& \textbf{Avg} & \conjcs{} & \disjcs{} & \negcs{} & \textbf{Avg} & \contraes{} & \distaes{} & \distbes{} \\
			\midrule
			Original                    & 0.47 & 0.44 & 0.61 & 0.37 & 0.89 & 0.79 & 0.93 & 0.94 \\
			w/o \textit{Unknown}  & 0.83 & 0.83 & 0.77 & 0.90 & 0.86 & 0.72 & 0.87 & 1.00 \\
			\bottomrule
		\end{tabular}%
	}
	\caption{\label{tab:results_statistical} Performance of \roberta{} checkpoint when finetuned on a subset of the training dataset without \textit{Unknown} label. We observe that the performance is still below the in-domain performances.}
	\vspace{-0.3cm}
\end{table}

\paragraph{Effect of Statistical Features}
In a contemporary work, \citet{sampling_algorithm} claims that deductive reasoning models inherently learn to use statistical features in the training data such as \#rules, \#facts, etc. Here, we demonstrate that it is not the complete reason for failure using the following control study. We finetune the \roberta{} model on a subset of the \texttt{All} dataset, where we restrict to two labels: \textit{True} and \textit{False}, by filtering out the \textit{Unknown} label. In our sampling algorithm, we ensure that each theory has the exact same number of \textit{True} and \textit{False} labeled statements in the training set. Thus, it is not possible to learn any statistic of the data, as the label distribution per theory is exactly uniform \footnote{\citet{sampling_algorithm} do not consider negations in the theory, and thus cannot ensure this property.}. Next, we evaluate the model on the \dataset{} evaluation sets, with the \textit{Unknown} label filtered in each set. The results are shown in Table \ref{tab:results_statistical}. We observe that, although the performance on this reduced test set is improved, there is still around 15\% gap on average with respect to performance on in-distribution data. This gap suggests that the model is not able to learn the logical operators robustly, even without any scope of spurious statistical features in the training set. Thus, this shows that although spurious correlation can lead to non-robust model behavior, it is not the sole reason for failure of these deductive reasoning models. This calls for the development of better inductive biases to teach the logical semantics more robustly to the language models.
\section{Related Works}

Reasoning in natural language has been a prevalent problem in NLP. There are multiple reasoning datasets, studying different aspects of reasoning over textual inputs. Natural Language Inference (NLI) \cite{nli_dataset} is a prominent dataset that requires reasoning over text to answer if a statement is entailed, contradicted, or neutral given a hypothesis. HotpotQA \citep{yang2018hotpotqa} tests multi-hop reasoning abilities that require comparisons and inferring missing bridge between sentences.
CLUTRR \citep{sinha-etal-2019-clutrr} tests whether models can infer biological relationships between entities in a context. RICA \cite{zhou-etal-2021-rica} requires the model to employ commonsense reasoning to answer questions based on a context.

Recently, there has been an increasing focus on evaluating the logical reasoning abilities of LMs. ReClor \cite{yu2020reclor} and LogiQA \cite{logiqa_dataset} are logical reasoning datasets derived from examinations. RuleTaker \citep{ruletaker} proposes synthetic deductive reasoning datasets that uses only the knowledge in the context. There are very limited works that probe the logical reasoning abilities of language models (LMs). FaiRR \cite{sanyal2022fairr} tests the robustness of logical reasoning models when the subjects and attributes in the context are altered to out-of-distribution terms. In a contemporary work, \citet{sampling_algorithm} show that language models can learn to use statistical features that can be present in deductive reasoning datasets. To the best of our knowledge, \dataset{} is the first dataset that tests how robust these LMs are to different \textit{logical} perturbations.
\section{Conclusion}
In this paper, we proposed \dataset{}, a diagnostic benchmark to test the logical robustness of deductive reasoning models. In \dataset{}, we propose two evaluation sets, \contrast{} and \equivalence{}, each probing different logical reasoning abilities. Overall, we find that fine-tuning LMs such as RoBERTa and T5 on deductive reasoning datasets is not sufficient to learn the semantics of the logical operators conjunction, disjunction, and negation. Although well-aligned training dataset improves model performance, the models still find it challenging to understand negations, both in \contrast{} and \equivalence{} sets. We demonstrate some interesting shortcoming of LMs designed for logical reasoning, that can eventually enable building better reasoning models.

\section{Limitation}
A key limitation of the work is the synthetic nature of the dataset. While it is ideal to explore more natural theories, it makes the systematic logical perturbation process very challenging. Thus, in this work, we resort to using synthetic datasets, but aim to bridge this gap in future works. Another limitation is the complexity of the datasets we explore. We use fairly simple logical rules and constructs for \dataset{}. Some more complex forms of logical reasoning-based theories can potentially reveal even more limitations of deductive reasoning models. Another interesting aspect we do not explore in this scope is potential techniques to improve these models on deductive reasoning tasks. This might involve trying different inductive biases in the form of architectural designs, more specialized datasets, etc.

\section*{Acknowledgments}
This research is supported in part by the Office of the Director of National Intelligence (ODNI), Intelligence Advanced Research Projects Activity (IARPA), via Contract No. 2019-19051600007, the DARPA MCS program under Contract No. N660011924033, the Defense Advanced Research Projects Agency with award W911NF-19-20271, NSF IIS 2048211, NSF SMA 1829268, and gift awards from Google, Amazon, JP Morgan and Sony. We would like to thank all the collaborators in USC INK research lab for their constructive feedback on the work.

\bibliography{custom}
\bibliographystyle{acl_natbib}

\clearpage

\appendix

\section{Model implementation details}
\label{app:implementation_details}
\begin{figure}[t]
	\centering
	\includegraphics[width=0.85\columnwidth]{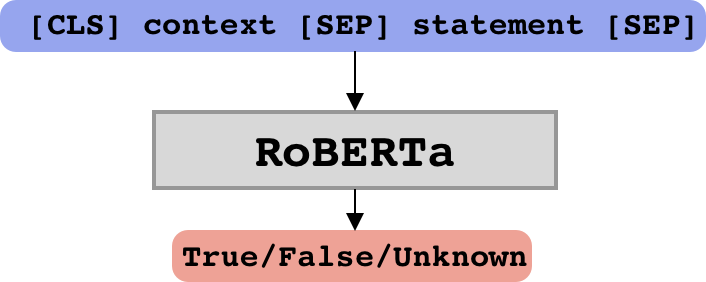}
	\caption{\textbf{Overview of the \roberta{} model} - The \textit{context} (containing the facts and rules) and the \textit{statement} are concatenated together as input and passed into a RoBERTa-Large model. The model is trained on cross entropy loss for a 3-class classification task.}
	\label{fig:model}
\end{figure}

In this section, we describe the implementation details of the language models used to evaluate \dataset{}.
\begin{itemize}
    \item \textbf{\roberta{}}: Following RuleTaker \cite{ruletaker}, we use a pre-trained \roberta{} \cite{liu2019roberta} model to perform the classification task. Specifically, we input in the format $[CLS]~T~[SEP]~s~[SEP]$ to the \roberta{} model, and extract the $[CLS]$ embedding to predict the label. The schematics of the RoBERTa model input is shown in Figure \ref{fig:model}. Here, $T$ is the theory which is the concatenation of the facts and rules, and $s$ is the statement. We use Cross Entropy loss to fine-tune the model on the training dataset.

    \item \textbf{T5-Large}: Following ProofWriter \cite{proofwriter}, we train a T5-Large \cite{raffel2019exploring} model for the deductive reasoning task. For this, we add a prefix to the T5-Large's input and generate the output in a fixed format. Specifically, we give the input in the format: $\$answer\$~;~\$question\$ = s~;~\$context\$ = T$. Here, $T$ is the theory which is the concatenation of the facts and rules, and $s$ is the statement. And the output is defined to be in format: $\$answer\$ = True/False/Unknown$. The model is trained on the default language modeling loss to match the output format. At evaluation time, we match the output template with the above description and generate the model's predicted label accordingly.
    
    \item \textbf{T5-3B}: Similar to T5-Large above, we use the T5-3B checkpoint.
    
    \item \textbf{T5-11B}: Similar to T5-Large above, we use the T5-11B checkpoint.
    
    \item \textbf{GPT3}: We use GPT-3 \cite{gpt3} for model evaluation to check its performance on our \conjcs{},\disjcs{},\negcs{}. Following \cite{prompts}, we experiment with all the prompts for the NLI task and select a prompt which performs the best on the evaluation datasets. We experiment with inserting 3 demonstrations and 10 demonstrations before the sentence and find that the performances are nearly same. So, we finally use the prompt named ``based on the previous passage'' which will give the input in a format as shown below:
    
    \textbf{Input Template:}\\
    \colorbox[RGB]{245,245,245}{\{\{premise\}\} Based on the previous passage,}\\
    \colorbox[RGB]{245,245,245}{is it true that "\{\{hypothesis\}\}"? Yes or no?}\\
    
    \textbf{Output Template:}\\
    \colorbox[RGB]{245,245,245}{\{\{ answer\_choices[label] \}\}}\\
    
    \textbf{Answer Choices Template:}\\
    \colorbox[RGB]{245,245,245}{Yes ||| Maybe ||| No}\\

    We use 3 demonstrations for each sample to limit the total tokens evaluated using the OpenAI GPT-3 API \footnote{\url{https://openai.com/api/}}.
\end{itemize}

\section{Hyperparameter Details}
\label{app:hyperparameters}
Here we use RoBERTa-Large \cite{liu2019roberta}and T5-Large,T5-3B,T5-11B \cite{raffel2019exploring} models for the 3-class deductive reasoning classification task. Only use GPT-3 \cite{gpt3} for evaluation. We train the pre-trained checkpoints available in the Hugging Face \cite{wolf-etal-2020-transformers} Transformers library. For \roberta{} model, we use AdamW \cite{loshchilov2018decoupled} with learning rate $1e\mhyphen5$. For T5-Large T5-3B and T5-11B, we use AdamW with learning rate $1e\mhyphen4$, adamw\_epsilon $1e\mhyphen6$, warmup\_ratio $1e\mhyphen1$, weight\_decay $1e\mhyphen2$. 
All these models are trained with batch size $8$ on Nvidia Quadro RTX 8000 GPUs. For RoBERTa-Large and T5-Large,training a single task on one GPU costs nearly 8 hours on average. For T5-3B and T5-11B, we use 4 GPUs to train the model and averagely need 5 and 10 hours for one epoch.

\section{Dataset Statistics}
\label{app:dataset_statistics}
In this section, we describe the training and evaluation dataset statistics. We first train the model on the datasets in Table \ref{tab:train_data_stats}. Each dataset comprises of different types of logical operators to help us in understanding the effect of different logical operators. Then we evaluate the trained models on evaluation datasets mentioned in Table \ref{tab:test_data_stats}. For evaluation, we test the model on two subsets of \dataset{}: \contrast{} set and \equivalence{} set. Each set is further sub-categorized into three different parts, based on the type of perturbations.

\begin{table}[t]
    \centering
    \resizebox{0.8\columnwidth}{!}{%
        \begin{tabular}{lccc}
            \toprule
            \textbf{Training Dataset}  & \textbf{Train} & \textbf{Dev} & \textbf{Test}   \\
            \midrule
           \texttt{NOT}           & 50000 & 10000 & 10000   \\
           \texttt{AND+NOT}      & 50000 & 10000 & 10000   \\
           \texttt{OR+NOT}       & 50000 & 10000 & 10000   \\
           \texttt{All}  & 50000 & 10000 & 10000   \\
            \bottomrule
        \end{tabular}
    }
    \caption{\label{tab:train_data_stats} Training dataset statistics. Please refer to Section \ref{app:dataset_statistics} for more details.}
\end{table}

\begin{table}[t]
    \centering
    \resizebox{0.7\columnwidth}{!}{%
        \begin{tabular}{lc}
            \toprule
            \textbf{Evaluation Set}  & \textbf{Number of instances}   \\
            \midrule
            \conjcs{}  & 20000   \\
            \disjcs{}  & 20000   \\
            \negcs{} & 20000   \\
            \midrule
            \contraes{}   & 20000  \\
            \distaes{}   & 20000   \\
            \distbes{}   & 20000   \\
            \bottomrule
        \end{tabular}
    }
    \caption{\label{tab:test_data_stats} Evaluation dataset statistics. Refer to Section \ref{app:dataset_statistics} for more details.}
\end{table}

\section{Filtering Statistical Features}
\label{app:statistical_features}
\begin{figure*}
	\centering
	\begin{subfigure}{.4\textwidth}
		\centering
		\includegraphics[width=\columnwidth]{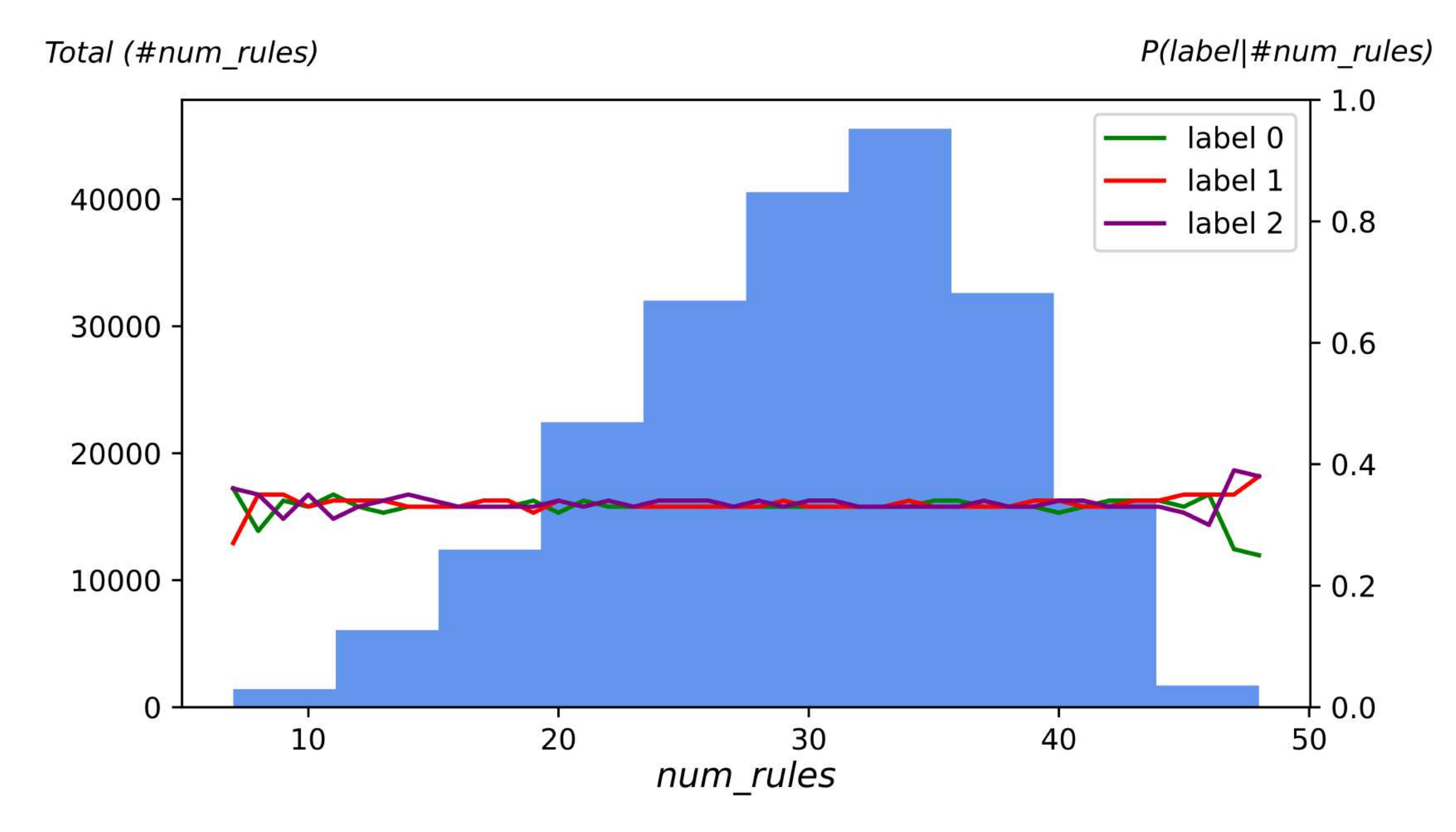}
		\caption{\# Rules}
	\end{subfigure}%
	\begin{subfigure}{.4\textwidth}
		\centering
		\includegraphics[width=\columnwidth]{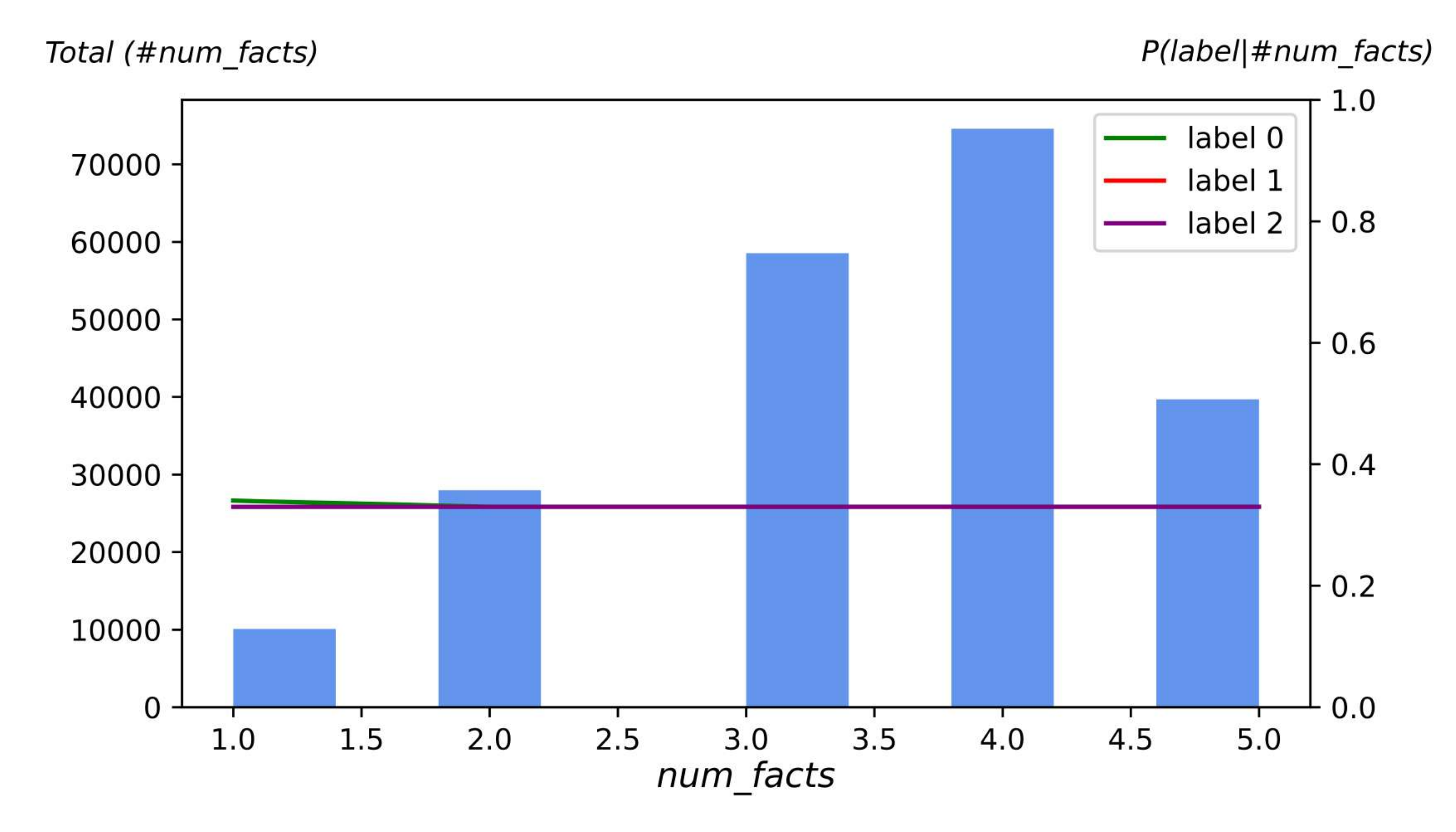}
		\caption{\# Facts}
	\end{subfigure}%
	\\
	\begin{subfigure}{.4\textwidth}
		\centering
		\includegraphics[width=\columnwidth]{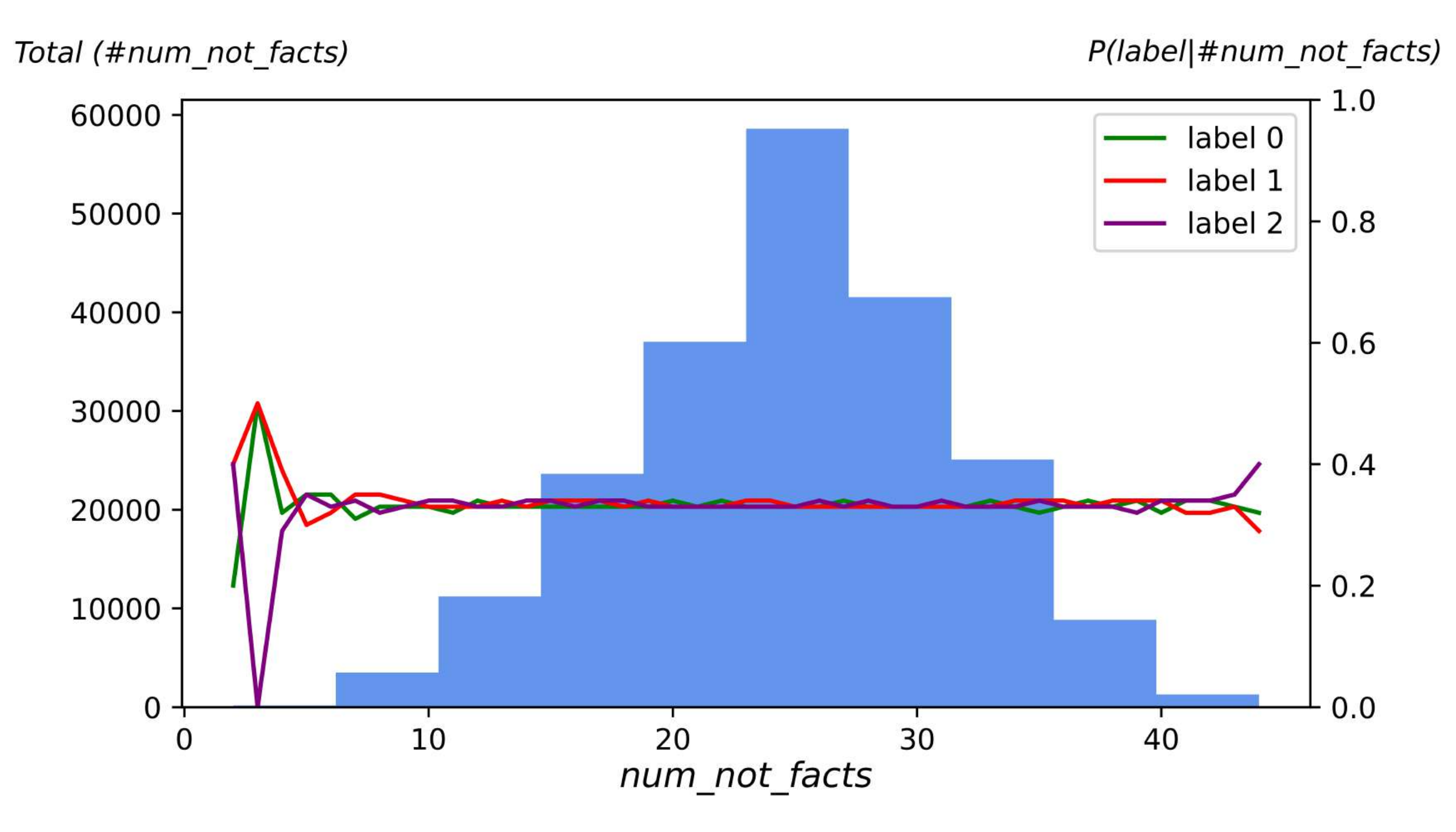}
		\caption{\# Facts with Negations}
	\end{subfigure}
	\begin{subfigure}{.4\textwidth}
		\centering
		\includegraphics[width=\columnwidth]{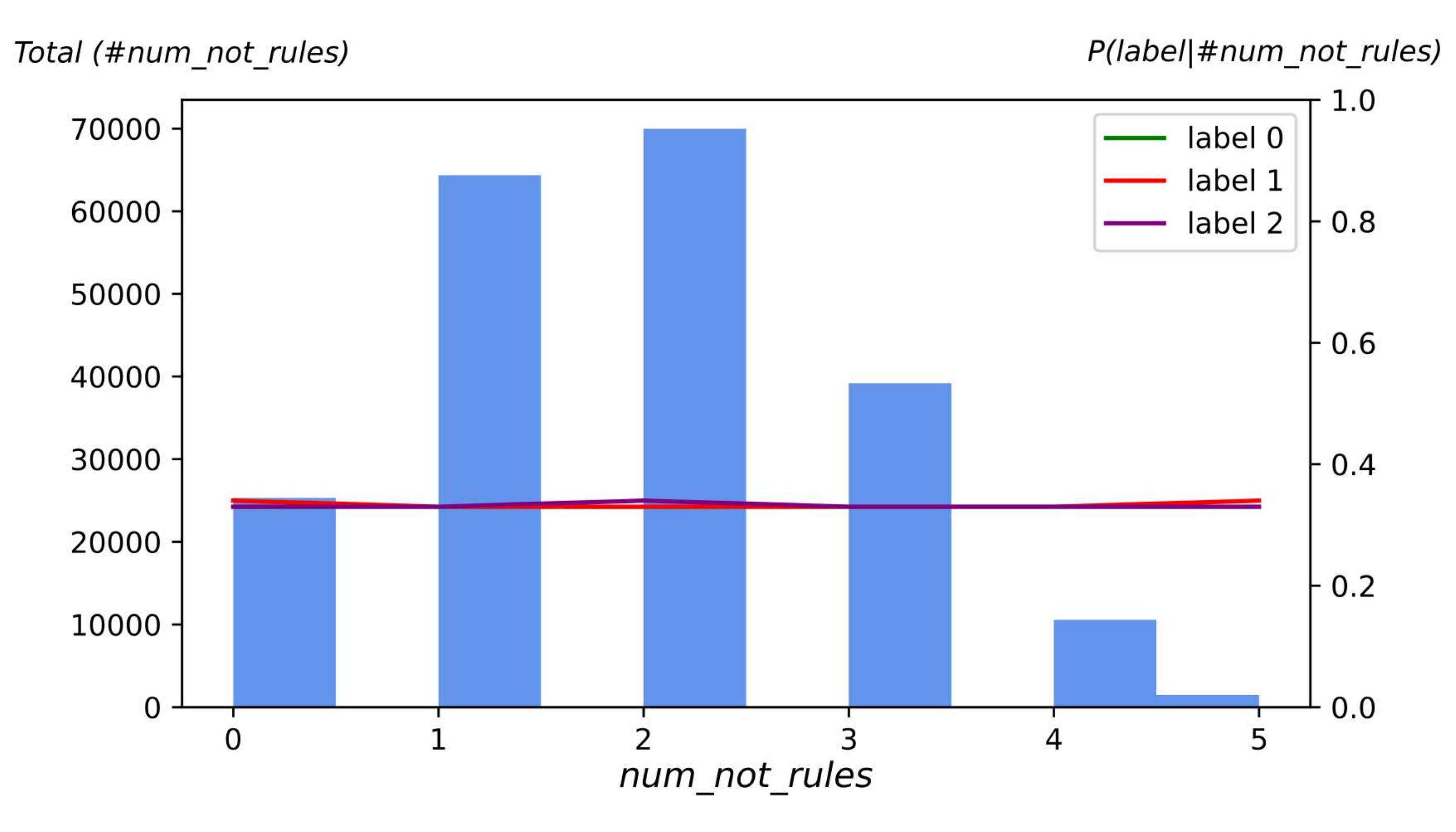}
		\caption{\# Rules with Negations}
	\end{subfigure}
	\begin{subfigure}{.4\textwidth}
		\centering
		\includegraphics[width=\columnwidth]{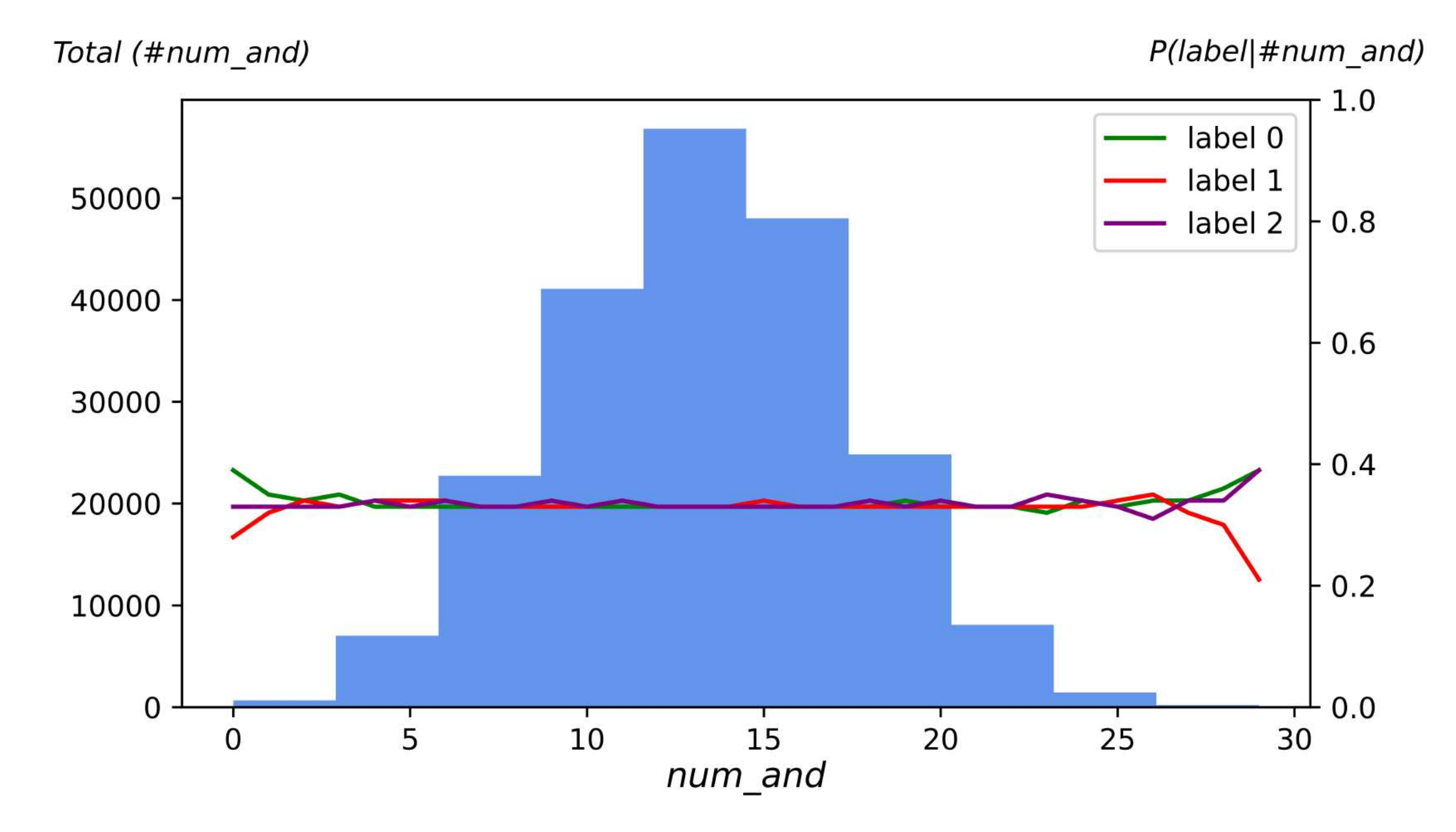}
		\caption{\# Rules with Conjunctions}
	\end{subfigure}
	\begin{subfigure}{.4\textwidth}
		\centering
		\includegraphics[width=\columnwidth]{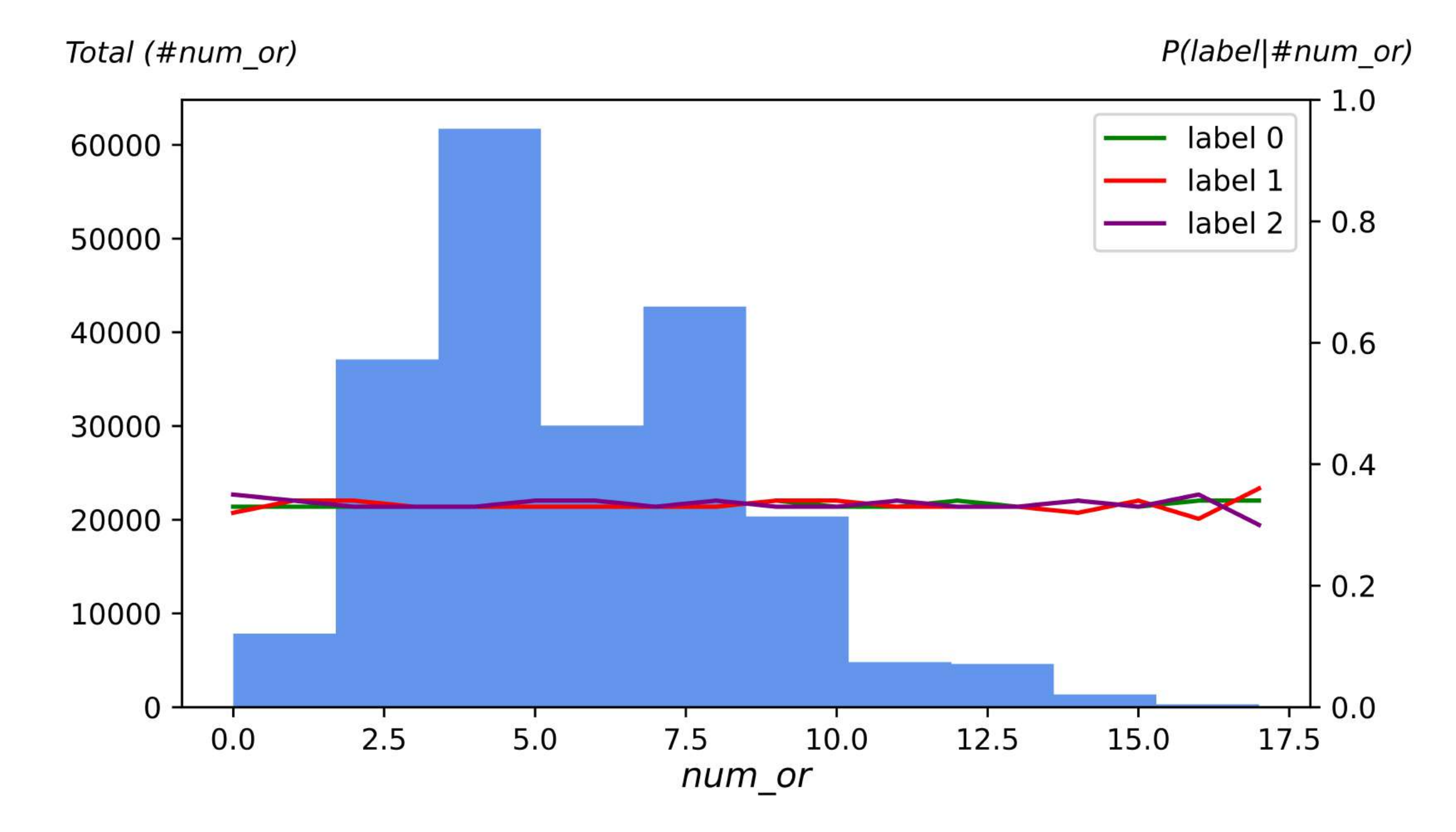}
		\caption{\# Rules with Disjunctions}
	\end{subfigure}
	\begin{subfigure}{.4\textwidth}
		\centering
		\includegraphics[width=\columnwidth]{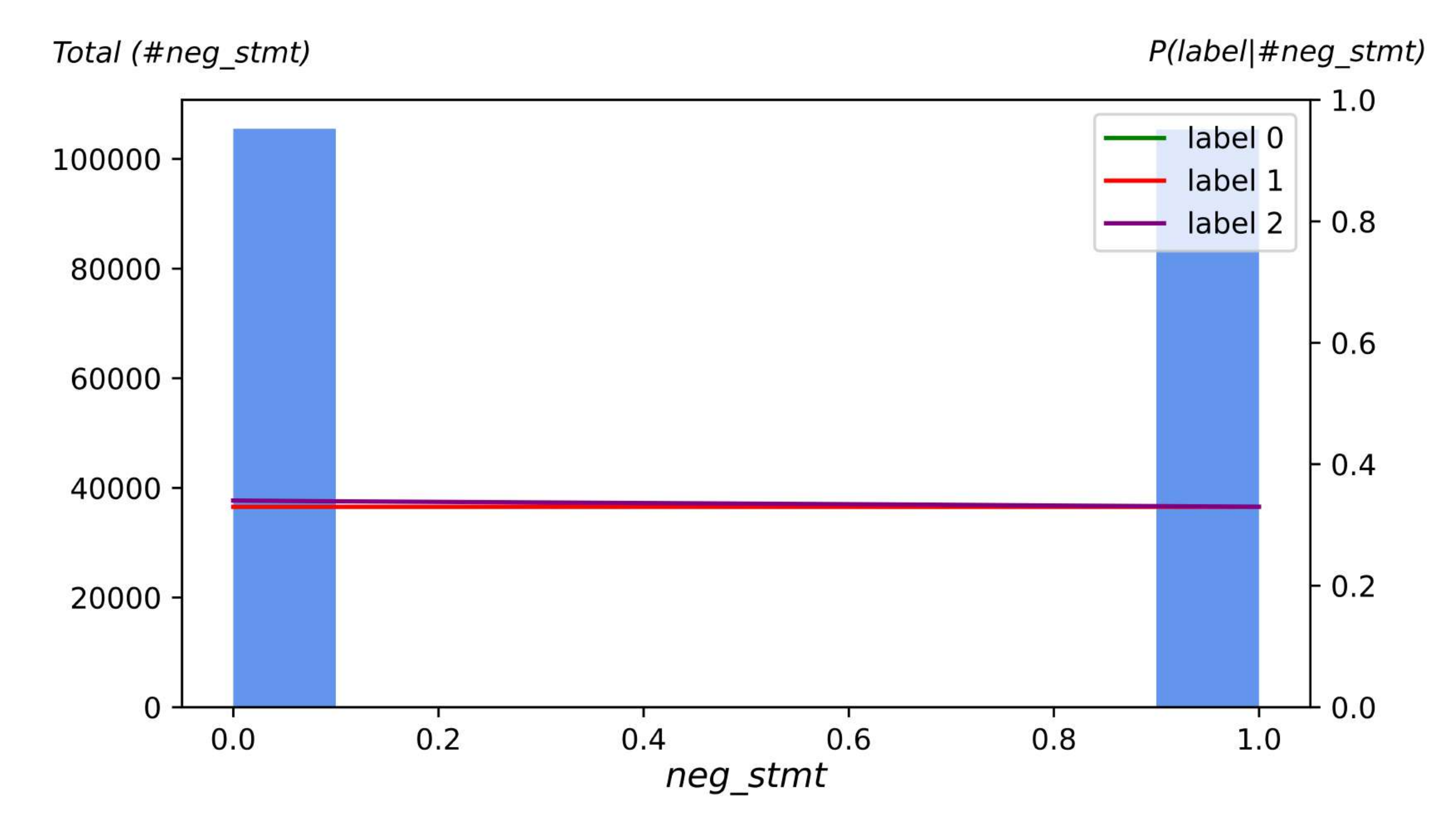}
		\caption{\# Statements with Negations}
	\end{subfigure}
	\caption{\label{fig:dataset_statistic_plots} Plots of label distribution with respect to different statistical features for the \texttt{All} dataset after our filtering techniques are used. Additionally, we plot the histogram of the count of instances for each feature value.}
\end{figure*}

In Figure \ref{fig:dataset_statistic_plots}, we show the plots of the label distribution for the following statistical features in the input theory and statement: \#rules, \#facts, \#facts with negation, \#rules with negation, \#rules with conjunction, \#rules with disjunction, and \#statements with negations. We observe that there is no significant bias between any of these features and the task label. Additionally, we show the count histogram of the instances in blue. Overall, our dataset filtering is able to remove some of the count-based statistical features.

\section{Contrast Perturbations}
\label{app:contrast_perturbations}
Following Section \ref{sec:contrast_sets}, we show the conjunction, disjunction, and negation contrast perturbations for the case when base theory's label is \textit{False} in Tables \ref{tab:conj_contrast_list_2}, \ref{tab:disj_contrast_list_2}, and \ref{tab:neg_contrast_list_2}, respectively.

For the conjunction contrast set perturbations in Table \ref{tab:conj_contrast_list} and \ref{tab:conj_contrast_list_2}, the first row is a base theory which is used to generate these contrast sets. In the next set of triads, the rule is modified to have an unseen predicate $t$ in conjunction with the existing rule LHS. Here $t$ is a predicate that is not part of the existing facts and inferences in the theory (hence, referred to as unseen predicate). Additionally, we add $t$ (or $\neg t$) as part of the facts in the theory. This lead to modification of the label as shown in rows 3-4. For the next set of triads, we modify the base rule to have a negated rule RHS $\neg q$. The corresponding label changes are shown in rows 5-7. In Table \ref{tab:conj_contrast_list_2}, we assume the label of the statement is \textit{False} for the base theory in row 1. Similar perturbations are possible for the label \textit{True}, and is shown in Table \ref{tab:conj_contrast_list} in \ref{sec:contrast_sets}. We group these perturbations into three classes as shown in Table \ref{tab:conj_contrast_list_2}: \textsc{base}, \textsc{conj}, \textsc{conj+neg}. These groups are based on which logical operator is the new addition with respect to the base theory. If a model performs accurately on this contrast set, we expect that the model understands the semantics of conjunction and negation logical operators reasonably well.

Similar to the \conjcs{} above, we show the perturbations considered in the \disjcs{} in Table \ref{tab:disj_contrast_list} and \ref{tab:disj_contrast_list_2}, where the distractor is added to the rule LHS using disjunction ($\lor$). Lastly, we show the \negcs{} perturbations in Tables \ref{tab:neg_contrast_list} and \ref{tab:neg_contrast_list_2}, where negations are added to the rule LHS and/or RHS.

\begin{table}[t]
	\centering
	\resizebox{0.85\columnwidth}{!}{%
		\begin{tabular}{lcccc}
			\toprule
			\textbf{Modified Rule} & \textbf{Facts}	& \textbf{Statement}	& \textbf{Label}    & \textbf{Group}	\\
			\midrule
			$p \implies q$ & $\{p\}$ & $q$  & \textit{True} & \textsc{base} \\
			\midrule
			$p \lor t \implies q$ & $\{p\}$ & $q$  & \textit{True}  & \textsc{disj} \\
			$p \lor t \implies q$ & $\{p, t\}$ & $q$  & \textit{True}   & \textsc{disj} \\
			\midrule
			$p \lor t \implies q$ & $\{\neg p, \neg t\}$ & $q$  & \textit{Unknown}  & \textsc{disj+neg} \\
			$p \lor t \implies \neg q$ & $\{p\}$ & $q$  & \textit{False}    & \textsc{disj+neg} \\
			$p \lor t \implies \neg q$ & $\{p, t\}$ & $q$  & \textit{False} & \textsc{disj+neg} \\
			$p \lor t \implies \neg q$ & $\{\neg p, \neg t\}$ & $q$  & \textit{Unknown} & \textsc{disj+neg} \\
			\bottomrule
		\end{tabular}%
	}
	\caption{\label{tab:disj_contrast_list} \small \textbf{Disjunction Contrast Perturbations}. The minimal edits done to a base theory (first row) for testing the disjunction and negation reasoning abilities. The group reflects the overall change in theory w.r.t. the base theory.}
\end{table}

\begin{table}[t]
	\centering
	\resizebox{0.82\columnwidth}{!}{%
		\begin{tabular}{lcccc}
			\toprule
			\textbf{Modified Rule} & \textbf{Facts}	& \textbf{Statement}	& \textbf{Label}    & \textbf{Group}	\\
			\midrule
			$p \implies q$ & $\{p\}$ & $q$  & \textit{True} & \textsc{base} \\
			\midrule
			$p \implies \neg q$ & $\{p\}$ & $q$  & \textit{False} & \textsc{neg} \\
			$\neg p \implies q$ & $\{p\}$ & $q$  & \textit{Unknown} & \textsc{neg} \\
			$\neg p \implies \neg q$ & $\{p\}$ & $q$  & \textit{Unknown} & \textsc{neg} \\
			\bottomrule
		\end{tabular}%
	}
	\caption{\label{tab:neg_contrast_list} \small \textbf{Negation Contrast Perturbations}. The minimal edits done to a base theory (first row) for testing the negation reasoning abilities.}
\end{table}

\begin{table}[t]
	\centering
	\resizebox{0.85\columnwidth}{!}{%
		\begin{tabular}{lcccc}
			\toprule
			\textbf{Modified Rule} & \textbf{Facts}	& \textbf{Statement}	& \textbf{Label}    & \textbf{Group}	\\
			\midrule
			$p \implies \neg q$ & $\{p\}$ & $q$  & \textit{False} & \textsc{base} \\
			\midrule
			$p \land t \implies \neg q$ & $\{p\}$ & $q$  & \textit{Unknown}  & \textsc{conj} \\
			$p \land t \implies \neg q$ & $\{p, t\}$ & $q$  & \textit{False}  & \textsc{conj} \\
			$p \land t \implies \neg q$ & $\{p, \neg t\}$ & $q$  & \textit{Unknown}  & \textsc{conj+neg} \\
			\midrule
			$p \land t \implies  q$ & $\{p\}$ & $q$  & \textit{Unknown} & \textsc{conj+neg} \\
			$p \land t \implies  q$ & $\{p, t\}$ & $q$  & \textit{True}    & \textsc{conj+neg} \\
			$p \land t \implies  q$ & $\{p, \neg t\}$ & $q$  & \textit{Unknown} & \textsc{conj+neg} \\
			\bottomrule
		\end{tabular}%
	}
	\caption{\label{tab:conj_contrast_list_2} \small \textbf{Conjunction Contrast Perturbations}. These are perturbations for testing conjunction and negation reasoning abilities. First row is the base theory being perturbed. Please refer to Appendix \ref{app:contrast_perturbations} for more details.}
\end{table}

\begin{table}[t]
	\centering
	\resizebox{0.85\columnwidth}{!}{%
		\begin{tabular}{lcccc}
			\toprule
			\textbf{Modified Rule} & \textbf{Facts}	& \textbf{Statement}	& \textbf{Label}    & \textbf{Group}	\\
			\midrule
			$p \implies q$ & $\{p\}$ & $q$  & \textit{False} & \textsc{base} \\
			\midrule
			$p \lor t \implies \neg q$ & $\{p\}$ & $q$  & \textit{False}  & \textsc{disj} \\
			$p \lor t \implies \neg q$ & $\{p, t\}$ & $q$  & \textit{False}   & \textsc{disj} \\
			$p \lor t \implies \neg q$ & $\{\neg p, \neg t\}$ & $q$  & \textit{Unknown}  & \textsc{disj+neg} \\
			\midrule
			$p \lor t \implies q$ & $\{p\}$ & $q$  & \textit{True}    & \textsc{disj+neg} \\
			$p \lor t \implies q$ & $\{p, t\}$ & $q$  & \textit{True} & \textsc{disj+neg} \\
			$p \lor t \implies q$ & $\{\neg p, \neg t\}$ & $q$  & \textit{Unknown} & \textsc{disj+neg} \\
			\bottomrule
		\end{tabular}%
	}
	\caption{\label{tab:disj_contrast_list_2} \small \textbf{Disjunction Contrast Perturbations}. These are perturbations for testing disjunction and negation reasoning abilities. First row is the base theory being perturbed. Please refer to Appendix \ref{app:contrast_perturbations} for more details.}
\end{table}

\begin{table}[t]
	\centering
	\resizebox{0.85\columnwidth}{!}{%
		\begin{tabular}{lcccc}
			\toprule
			\textbf{Modified Rule} & \textbf{Facts}	& \textbf{Statement}	& \textbf{Label}    & \textbf{Group}	\\
			\midrule
			$p \implies q$ & $\{p\}$ & $q$  & \textit{False} & \textsc{base} \\
			\midrule
			$p \implies \neg q$ & $\{p\}$ & $q$  & \textit{True} & \textsc{neg} \\
			$\neg p \implies q$ & $\{p\}$ & $q$  & \textit{Unknown} & \textsc{neg} \\
			$\neg p \implies \neg q$ & $\{p\}$ & $q$  & \textit{Unknown} & \textsc{neg} \\
			\bottomrule
		\end{tabular}%
	}
	\caption{\label{tab:neg_contrast_list_2} \small \textbf{Negation Contrast Perturbations}. The minimal edits done to a base theory (first row) for testing the negation reasoning abilities.}
\end{table}

\section{\contrast{} set breakdown}
\label{app:contrast_set_breakdown}
In this section, we further discuss the performance of the LMs on each group of the \contrast{} set. From Tables \ref{tab:contrast_breakdown_roberta}, \ref{tab:contrast_breakdown_t5} and \ref{tab:contrast_breakdown_t5_3b} we can say that the models generally perform worse when they need to handle more complicated compound rules (\textsc{Conj + Neg} $>$ \textsc{Conj} $>$ \textsc{Base} (where $>$ means harder)). Additionally, we find that when we add more compound rules in the training dataset, the performance is generally better. Giving more complex rules can lead to further drops in performance, as noted by performance on the \textsc{Conj+Neg} and \textsc{Disj+Neg}. Models trained on the dataset with aligned operators instead of \texttt{All} dataset is better e.g., model trained on \texttt{AND+NOT} get best result at \textsc{Conj+Neg}.

It is easy to see that model with larger amount of parameters give more consistent and better result at \conjcs{}, \disjcs{}, \negcs{} which means the model learned more semantics of logic from language and is more robust.

\begin{table*}[t]
	\centering
	\resizebox{0.88\textwidth}{!}{%
		\begin{tabular}{lcccccccc}
			\toprule
			\multirow{3}{*}{\textbf{\contrast{} set breakdown}}		& \multicolumn{3}{c}{\conjcs{}}	& \multicolumn{3}{c}{\disjcs{}} & \multicolumn{2}{c}{\negcs{}}\\
			\cmidrule(r){2-4} \cmidrule(r){5-7} \cmidrule(r){8-9}
			& \textsc{Base}	& \textsc{Conj}	& \textsc{Conj + Neg}	& \textsc{Base}	& \textsc{Disj}	& \textsc{Disj + Neg}  &
			\textsc{Base}	& \textsc{Neg} \\
			\midrule
			\texttt{NOT}& 1.00 &	0.28 &	0.28 &	1.00 &	0.62 &	0.28 &	1.00 &	0.22 \\
			\texttt{AND+NOT}& 1.00 &	0.50 &	0.48 &	1.00 &	0.56 &	0.43 &	1.00 &	0.37  \\
			\texttt{OR+NOT} & 1.00 &	0.32 &	0.29 &	1.00 &	0.74 &	0.52 &	1.00 &	0.31 \\
			\texttt{All}& 1.00 &	0.41 &	0.34 &	1.00 &	0.81 &	0.50 &	1.00 &	0.26 \\
			\bottomrule
		\end{tabular}%
	}
	\caption{\label{tab:contrast_breakdown_roberta} Performance breakdown of \roberta{} with different groups of \contrast{} set. Please refer to Appendix \ref{app:contrast_set_breakdown} for more details.}
\end{table*}

\begin{table*}[t]
	\centering
	\resizebox{0.88\textwidth}{!}{%
		\begin{tabular}{lcccccccc}
			\toprule
			\multirow{3}{*}{\textbf{\contrast{} set breakdown}}		& \multicolumn{3}{c}{\conjcs{}}	& \multicolumn{3}{c}{\disjcs{}} & \multicolumn{2}{c}{\negcs{}}\\
			\cmidrule(r){2-4} \cmidrule(r){5-7} \cmidrule(r){8-9}
			& \textsc{Base}	& \textsc{Conj}	& \textsc{Conj + Neg}	& \textsc{Base}	& \textsc{Disj}	& \textsc{Disj + Neg}  &
			\textsc{Base}	& \textsc{Neg} \\
			\midrule
			\texttt{NOT}& 1.00 &	0.30 &	0.23 &	1.00 &	0.68 &	0.43 &	1.00 &	0.30 \\
			\texttt{AND+NOT}& 1.00 &	0.40 &	0.31 &	1.00 &	0.69 &	0.43 &	1.00 &	0.32\\
			\texttt{OR+NOT} & 1.00 &	0.31 &	0.23 &	1.00 &	0.77 &	0.49 &	1.00 &	0.32\\
			\texttt{All} & 1.00 &	0.34 &	0.24 &	1.00 &	0.84 &	0.48 &	1.00 &	0.30 \\
			\bottomrule
		\end{tabular}%
	}
	\caption{\label{tab:contrast_breakdown_t5} Performance breakdown of T5-Large with different groups of \contrast{} set. Please refer to Appendix \ref{app:contrast_set_breakdown} for more details.}
\end{table*}

\begin{table*}[t]
	\centering
	\resizebox{0.88\textwidth}{!}{%
		\begin{tabular}{lcccccccc}
			\toprule
			\multirow{3}{*}{\textbf{\contrast{} set breakdown}}		& \multicolumn{3}{c}{\conjcs{}}	& \multicolumn{3}{c}{\disjcs{}} & \multicolumn{2}{c}{\negcs{}}\\
			\cmidrule(r){2-4} \cmidrule(r){5-7} \cmidrule(r){8-9}
			& \textsc{Base}	& \textsc{Conj}	& \textsc{Conj + Neg}	& \textsc{Base}	& \textsc{Disj}	& \textsc{Disj + Neg}  &
			\textsc{Base}	& \textsc{Neg} \\
			\midrule
			\texttt{NOT}& 0.96 &	0.56 &	0.53 &	0.96 &	0.61 &	0.57 &	0.96 &	0.55\\
			\texttt{AND+NOT}& 0.91 &	0.58 &	0.53 &	0.90 &	0.62 &	0.54 &	0.90 &	0.51\\
			\texttt{OR+NOT} & 0.95 &	0.37 &	0.39 &	0.93 &	0.74 &	0.65 &	0.93 &	0.53\\
			\texttt{All} & 0.93 &	0.54 &	0.49 &	0.93 &	0.75 &	0.62 &	0.94 &	0.49 \\
			\bottomrule
		\end{tabular}%
	}
	\caption{\label{tab:contrast_breakdown_t5_3b} Performance breakdown of T5-3B with different groups of \contrast{} set. Please refer to Appendix \ref{app:contrast_set_breakdown} for more details.}
\end{table*}

\section{Result breakdown by label}
\label{app:results_by_label}

We report the performance for each label in Tables \ref{tab:results_contrast_by_label_roberta} to \ref{tab:results_equivalence_by_label_t5_3b}, for both the \contrast{} and \equivalence{} sets. We find that T5-3B model, the largest model among the three models, get a good result for \textit{Unknown} while other two models are not good at it. It shows that models with large amount of parameters can better learn to predict the \textit{Unknown} label, which is relatively harder than the other two labels. Also, we find that \equivalence{} set is an easier task in general than \contrast{} set and the performances are stable across three models.

\begin{table*}[t]
	\centering

	\resizebox{0.88\textwidth}{!}{%
		\begin{tabular}{lccccccccc}
			\toprule
			\multirow{3}{*}{\textbf{Training Dataset}} &
			\multicolumn{3}{c}{\conjcs{}} &
			\multicolumn{3}{c}{\disjcs{}} &
			\multicolumn{3}{c}{\negcs{}}\\
			\cmidrule(r){2-4} \cmidrule(r){5-7} \cmidrule(r){8-10} 
			& \textit{False} & \textit{True} & \textit{Unknown} & \textit{False} & \textit{True} & \textit{Unknown} &
			\textit{False} & \textit{True} & \textit{Unknown}\\
			\midrule
			\texttt{NOT}& 0.77 & 	0.76 &	0.24 &	0.68 &	0.66 &	0.14 &	0.73 &	0.74 &	0.18\\
			\texttt{AND+NOT} & 0.89 & 	0.87 &	0.48 &	0.71 &	0.71 &	0.38 &	0.67 &	0.69 &	0.49 \\
			\texttt{OR+NOT}& 0.93 & 	0.94 &	0.19 &	0.90 &	0.92 &	0.19 &	0.73 &	0.72 &	0.33 \\
			\texttt{All} & 0.88 & 	0.89 &	0.28 &	0.91 &	0.93 &	0.14 &	0.76 &	0.76 &	0.23 \\
			\bottomrule
		\end{tabular}%
	}
	\caption{\label{tab:results_contrast_by_label_roberta} Performance breakdown of \roberta{} with different labels for \contrast{} set. Please refer to Appendix \ref{app:results_by_label} for more details.}
\end{table*}

\begin{table*}[t]
	\centering
	\resizebox{0.88\textwidth}{!}{%
		\begin{tabular}{lccccccccc}
			\toprule
			\multirow{2}{*}{\textbf{Training Dataset}}		& \multicolumn{3}{c}{\textbf{Contrapositive}}	& \multicolumn{3}{c}{\textbf{Distributive 1}}	& \multicolumn{3}{c}{\textbf{Distributive 2}}   \\
			\cmidrule(r){2-4} \cmidrule(r){5-7} \cmidrule(r){8-10}
			& \textit{False} & \textit{True} & \textit{Unknown} & \textit{False} & \textit{True} & \textit{Unknown} & \textit{False} & \textit{True} & \textit{Unknown} \\
			\midrule
			\texttt{NOT}& 0.72 &	0.72 &	0.92 &	0.91 &	0.90 &	- & 0.73 &	0.75 &	- \\
			\texttt{AND+NOT} & 0.73 &	0.74 &	0.92 &	0.88 &	0.89 &	- & 0.94 &	0.94 &	-\\
			\texttt{OR+NOT}  & 0.72 &	0.73 &	0.89 &	0.85 &	0.89 &	- & 0.95 &	0.94 &	-\\
			\texttt{All} & 0.73 &	0.75 &	0.90 &	0.91 &	0.94 &	- & 0.93 &	0.94 &	- \\
			\bottomrule
		\end{tabular}%
	}
	\caption{\label{tab:results_equivalence_by_label_t5} Performance breakdown of \roberta{} with different labels for \equivalence{} set. Please refer to Appendix \ref{app:results_by_label} for more details.}
\end{table*}

\begin{table*}[t]
	\centering
	\resizebox{0.88\textwidth}{!}{%
		\begin{tabular}{lccccccccc}
			\toprule
			\multirow{3}{*}{\textbf{Training Dataset}} &
			\multicolumn{3}{c}{\conjcs{}} &
			\multicolumn{3}{c}{\disjcs{}} &
			\multicolumn{3}{c}{\negcs{}}\\
			\cmidrule(r){2-4} \cmidrule(r){5-7} \cmidrule(r){8-10} 
			& \textit{False} & \textit{True} & \textit{Unknown} & \textit{False} & \textit{True} & \textit{Unknown} &
			\textit{False} & \textit{True} & \textit{Unknown}\\
			\midrule
			\texttt{NOT}&	0.91 &	0.90 &	0.13 &	0.85 &	0.85 &	0.10 &	0.91 &	0.90 &	0.15 \\
			\texttt{AND+NOT} &	0.95 &	0.93 &	0.23 &	0.87 &	0.83 &	0.11 &	0.89 &	0.88 &	0.20 \\
			\texttt{OR+NOT}&	0.94 &	0.95 &	0.12 &	0.93 &	0.94 &	0.10 &	0.83 &	0.86 &	0.24 \\
			\texttt{All} &	0.95 &	0.93 &	0.13 &	0.95 &	0.94 &	0.07 &	0.92 &	0.90 &	0.14  \\
			\bottomrule
		\end{tabular}%
	}
	\caption{\label{tab:results_contrast_by_label_t5} Performance breakdown of T5-Large with different labels for \contrast{} set. Please refer to Appendix \ref{app:results_by_label} for more details.}
\end{table*}

\begin{table*}[t]
	\centering
	\resizebox{0.88\textwidth}{!}{%
		\begin{tabular}{lccccccccc}
			\toprule
			\multirow{2}{*}{\textbf{Training Dataset}}		& \multicolumn{3}{c}{\textbf{Contrapositive}}	& \multicolumn{3}{c}{\textbf{Distributive 1}}	& \multicolumn{3}{c}{\textbf{Distributive 2}}   \\
			\cmidrule(r){2-4} \cmidrule(r){5-7} \cmidrule(r){8-10}
			& \textit{False} & \textit{True} & \textit{Unknown} & \textit{False} & \textit{True} & \textit{Unknown} & \textit{False} & \textit{True} & \textit{Unknown} \\
			\midrule
			\texttt{NOT}& 0.72 &	0.72 &	0.83 &	0.90 &	0.90 &	- &	0.97 &	0.97 &	- \\
			\texttt{AND+NOT} & 0.72 &	0.72 &	0.86 &	0.92 &	0.88 &	- &	0.96 &	0.97 &	-\\
			\texttt{OR+NOT}    & 0.71 &	0.72 &	0.87 &	0.85 &	0.86 &	- &	0.97 &	0.98 &	- \\
			\texttt{All}  & 0.73 &	0.72 &	0.85 &	0.94 &	0.91 &	- &	0.98 &	0.98 &	- \\
			\bottomrule
		\end{tabular}%
	}
	\caption{\label{tab:results_equivalence_by_label_t5} Performance breakdown of T5-Large with different labels for \equivalence{} set. Please refer to Appendix \ref{app:results_by_label} for more details.}
\end{table*}

\begin{table*}[t]
	\centering
	\resizebox{0.88\textwidth}{!}{%
		\begin{tabular}{lccccccccc}
			\toprule
			\multirow{3}{*}{\textbf{Training Dataset}} &
			\multicolumn{3}{c}{\conjcs{}} &
			\multicolumn{3}{c}{\disjcs{}} &
			\multicolumn{3}{c}{\negcs{}}\\
			\cmidrule(r){2-4} \cmidrule(r){5-7} \cmidrule(r){8-10} 
			& \textit{False} & \textit{True} & \textit{Unknown} & \textit{False} & \textit{True} & \textit{Unknown} &
			\textit{False} & \textit{True} & \textit{Unknown}\\
			\midrule
			\texttt{NOT}& 0.93 &	0.94 &	0.53 &	0.79 &	0.82 &	0.43 &	0.91 &	0.94 &	0.46\\
			\texttt{AND+NOT} & 0.91 &	0.86 &	0.53 &	0.80 &	0.76 &	0.41 &	0.90 &	0.84 &	0.44\\
			\texttt{OR+NOT}& 0.95 &	0.95 &	0.30 &	0.92 &	0.92 &	0.37 &	0.93 &	0.93 &	0.41 \\
			\texttt{All}& 0.91 &	0.94 &	0.43 &	0.91 &	0.93 &	0.30 &	0.93 &	0.96 &	0.33 \\
			\bottomrule
		\end{tabular}%
	}
	\caption{\label{tab:results_contrast_by_label_t5_3b} Performance breakdown of T5-3B with different labels for \contrast{} set. Please refer to Appendix \ref{app:results_by_label} for more details.}
\end{table*}

\begin{table*}[t]
	\centering
	\resizebox{0.88\textwidth}{!}{%
		\begin{tabular}{lccccccccc}
			\toprule
			\multirow{2}{*}{\textbf{Training Dataset}}		& \multicolumn{3}{c}{\textbf{Contrapositive}}	& \multicolumn{3}{c}{\textbf{Distributive 1}}	& \multicolumn{3}{c}{\textbf{Distributive 2}}   \\
			\cmidrule(r){2-4} \cmidrule(r){5-7} \cmidrule(r){8-10}
			& \textit{False} & \textit{True} & \textit{Unknown} & \textit{False} & \textit{True} & \textit{Unknown} & \textit{False} & \textit{True} & \textit{Unknown} \\
			\midrule
			\texttt{NOT}& 0.71 &	0.72 &	0.92 &	0.77 &	0.81 &	- &	0.95 &	0.96 &	- \\
			\texttt{AND+NOT}  & 0.71 &	0.69 &	0.88 &	0.83 &	0.78 &	- &	0.95 &	0.90 &	-\\
			\texttt{OR+NOT}  & 0.70 &	0.70 &	0.92 &	0.81 &	0.82 &	- &	0.98 &	0.98 &	- \\
			\texttt{All} & 0.70 &	0.69 &	0.76 &	0.84 &	0.81 &	- &	0.99 &	0.98 &	- \\
			\bottomrule
		\end{tabular}%
	}
	\caption{\label{tab:results_equivalence_by_label_t5_3b} Performance breakdown of T5-3B with different labels for \equivalence{} set. Please refer to Appendix \ref{app:results_by_label} for more details.}
\end{table*}

\section{Human Evaluation}
\label{app:iaa_result}
We recruit three Computer Science graduates to annotate the datasets. To keep the annotation realistic, we sample 30 instances from each test subset of \dataset{} and ask the annotators to mark a label from \textit{True}, \textit{False}, and \textit{Unknown}. The question asked is: ``Does the theory entail or contradict the statement, or we cannot say anything about it?''. Overall, we find the average inter-annotator agreement to be around $0.79$, evaluated using the Fleiss' kappa score \footnote{\url{https://www.statsmodels.org/dev/generated/statsmodels.stats.inter_rater.fleiss_kappa.html}}.

\begin{algorithm}[t]
	\small
	\SetKwFunction{sample}{\textsc{Sample}}
	\SetKwFunction{validate}{\textsc{Validate}}
	\SetKwInOut{KwIn}{Input}
	\SetKwInOut{KwOut}{Output}

	\KwIn{$vocab$ containing the corpus of all predicates, $ruleset$ containing the set of valid rules, predicate negation probability $n_1$, statement negation probability $n_2$, max reasoning depth $d$.}
	\KwOut{A theory containing a set of $facts$ and $rules$, a $statement$, and a corresponding $label \in \{0, 1, 2\}$}

	$pred\_num \sim U[10, 30]$\\
	$preds \leftarrow \sample(vocab, pred\_num)$\\
	$\textrm{set}~ l \sim U[1, d] ~\textrm{and group} ~preds~ \textrm{into}~ l ~\textrm{layers}$\\
	$rules \leftarrow [~]$

	\For{$\textrm{predicate}~p~\textrm{in layer}~ 1 \leq i \leq l$}{
		$\textrm{Negate}~p~\textrm{with probability}~n_1$\\
		$q \sim U[0, 1]$\\
		$\textrm{assign label}~q~\textrm{to predicate}~p$\\
		\If{$i \ge 1$}{
			$k \sim U[1, 2]$\\
			$cand \leftarrow p~\textrm{in layer}~i-1~\textrm{with label}~q$\\
			$body \leftarrow \sample(cand, k)$\\
			\If{$len(body) > 1$}{
				$operator \leftarrow \sample([\land, \lor], 1)$\\
				$\textrm{Compose the predicates in the}~body~$\\
				$\textrm{using}~operator$\\
			}
			$r \leftarrow (body \implies p)$\\
			\eIf{$\validate(ruleset, r)$}{
				$\textrm{add}~r~\textrm{to}~rules$\\
			}{
				\tcc{\small Rule $r$ does not match any valid rule forms, so the predicate is not provable}
				$\textrm{assign label}~0~\textrm{to predicate}~p$\\
			}
		}
	}
	$facts \leftarrow \textrm{predicates in layer 1 with label 1}$\\
	$statement \leftarrow \sample(preds, 1)$\\
	$label \leftarrow~\textrm{pre-assigned label for}~statement$\\
	\If{$label == 1$}{
		$\textrm{Negate the}~statement~\textrm{with probability}~n_2$\\
		$label \leftarrow 2$\\
	}

	\KwRet{$(facts, rules, statement, label)$}
	\caption{\label{algo:sampling}Sampling Algorithm}
\end{algorithm}

\end{document}